\pdfoutput=1
\documentclass[11pt]{article}

\usepackage{amsmath,amsfonts,bm}

\def\eqref#1{equation~\ref{#1}}
\def\1{\bm{1}}

\DeclareMathAlphabet{\mathsfit}{\encodingdefault}{\sfdefault}{m}{sl}
\SetMathAlphabet{\mathsfit}{bold}{\encodingdefault}{\sfdefault}{bx}{n}

\DeclareMathOperator*{\argmax}{arg\,max}

\usepackage[preprint]{acl}

\usepackage{times}
\usepackage{latexsym}

\usepackage[T1]{fontenc}
\usepackage[utf8]{inputenc}

\usepackage{microtype}

\usepackage{inconsolata}

\usepackage{graphicx}

\usepackage{hyperref}
\usepackage{url}
\usepackage{enumitem}
\usepackage{algorithm,algpseudocode,algorithmicx}
\usepackage[capitalize]{cleveref}
\usepackage{colortbl}
\usepackage{booktabs}
\usepackage{wrapfig}
\usepackage{subcaption}
\usepackage{xcolor}
\usepackage{xspace}
\usepackage{trimclip}
\usepackage{stmaryrd}
\usepackage[normalem]{ulem}
\usepackage{tabularx}

\definecolor{mygreen}{RGB}{234,234,222}
\newcommand{\optima}{\textsc{Optima}\xspace}

\title{\includegraphics[width=1.5em]{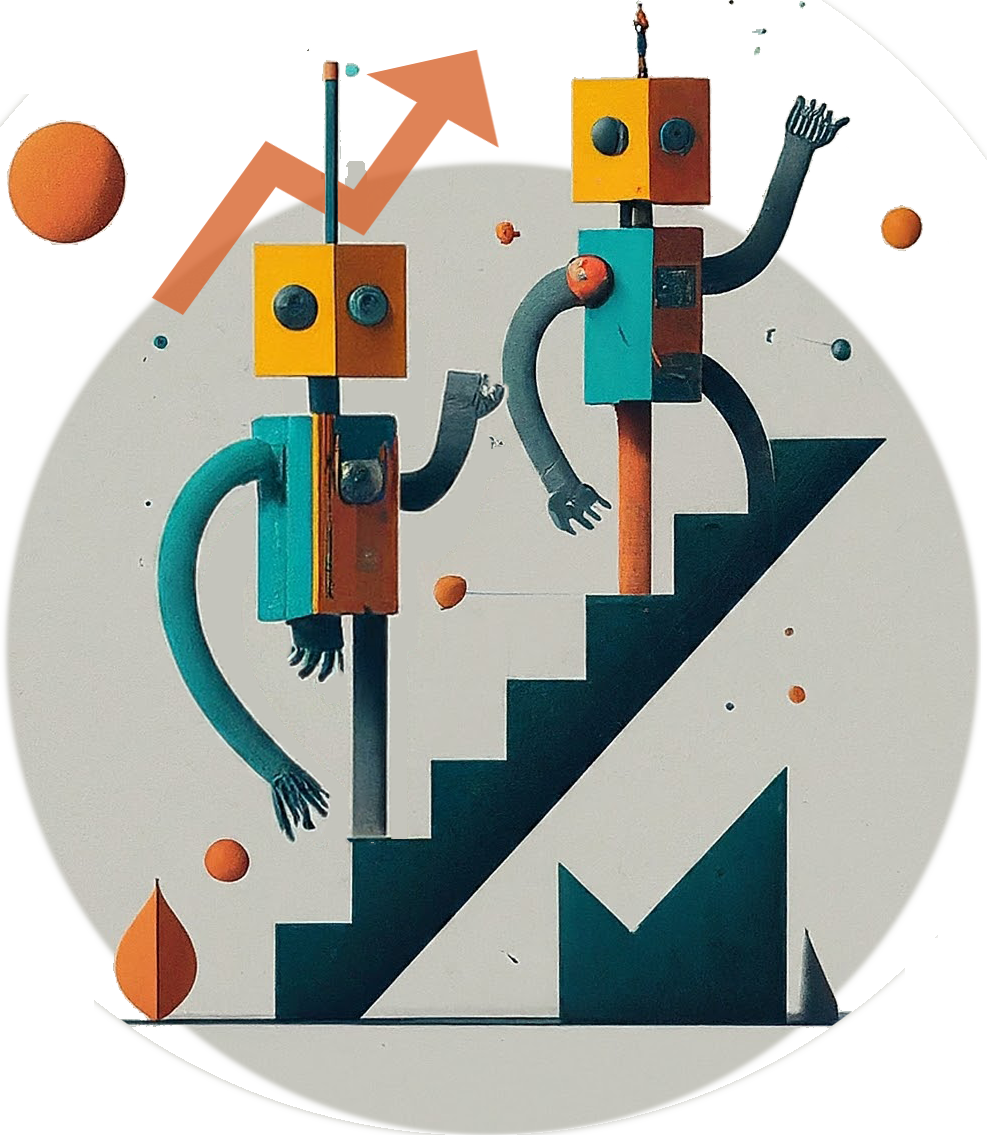}\textsc{Optima}: \underline{Opti}mizing Effectiveness and Efficiency for LLM-Based \underline{M}ulti-\underline{A}gent System}

\author{Weize Chen$^1\thanks{Equal Contribution.}$, Jiarui Yuan$^{1*}$, Chen Qian$^1$,\textbf{Cheng Yang$^2$, Zhiyuan Liu$^1$, Maosong Sun$^1$}\\
$^1$ Tsinghua University,
$^2$ Beijing University of Posts and Telecommunications\\
\texttt{\{chenwz21,yuanjr22\}@mails.tsinghua.edu.cn}, 
\texttt{liuzy@tsinghua.edu.cn}
}

\begin{document}
\maketitle
\begin{abstract}
Large Language Model (LLM) based multi-agent systems (MAS) show remarkable potential in collaborative problem-solving, yet they still face critical challenges: low communication efficiency, poor scalability, and a lack of effective parameter-updating optimization methods. We present \textbf{\optima}, a novel framework that addresses these issues by significantly enhancing \textit{both} communication efficiency and task effectiveness in LLM-based MAS through training. \optima employs an \textit{iterative generate, rank, select, and train} paradigm with a reward function balancing task performance, token efficiency, and communication readability. We explore various algorithms, including Supervised Fine-Tuning, Direct Preference Optimization, and their hybrid approaches, providing insights into their effectiveness-efficiency trade-offs. We integrate Monte Carlo Tree Search-inspired techniques for DPO data generation, treating conversation turns as tree nodes to explore diverse interaction paths. Evaluated on common multi-agent tasks, including information-asymmetric question answering and complex reasoning, \optima shows consistent and substantial improvements over single-agent baselines and vanilla MAS based on Llama 3 8B / 3.2 3B, achieving up to \textit{2.8x performance gain with less than 10\% tokens} on tasks requiring heavy information exchange. Moreover, \optima's efficiency gains enable more effective compute utilization during inference, leading to improved inference-time scaling laws. By addressing fundamental challenges in LLM-based MAS, \optima shows the potential towards scalable, efficient, and effective MAS.
\end{abstract}

\begin{figure}[t!]
    \centering
    % \vspace{-1em}
    \begin{subfigure}[b]{0.48\linewidth}
        \includegraphics[width=\linewidth]{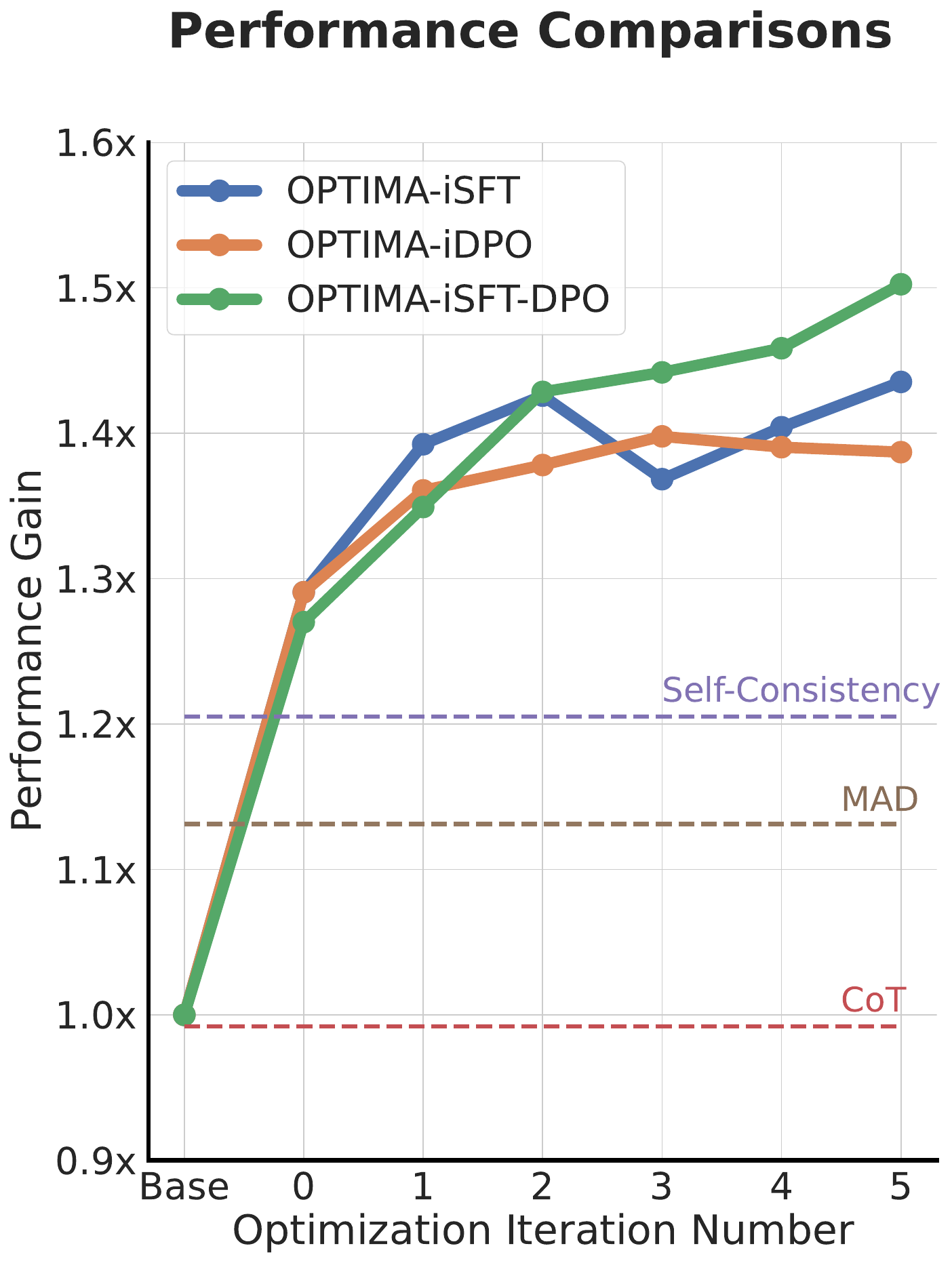}
        % \caption{Caption}
        \label{fig:performance-overview-perf}
    \end{subfigure}
    \begin{subfigure}[b]{0.48\linewidth}
        \includegraphics[width=\linewidth]{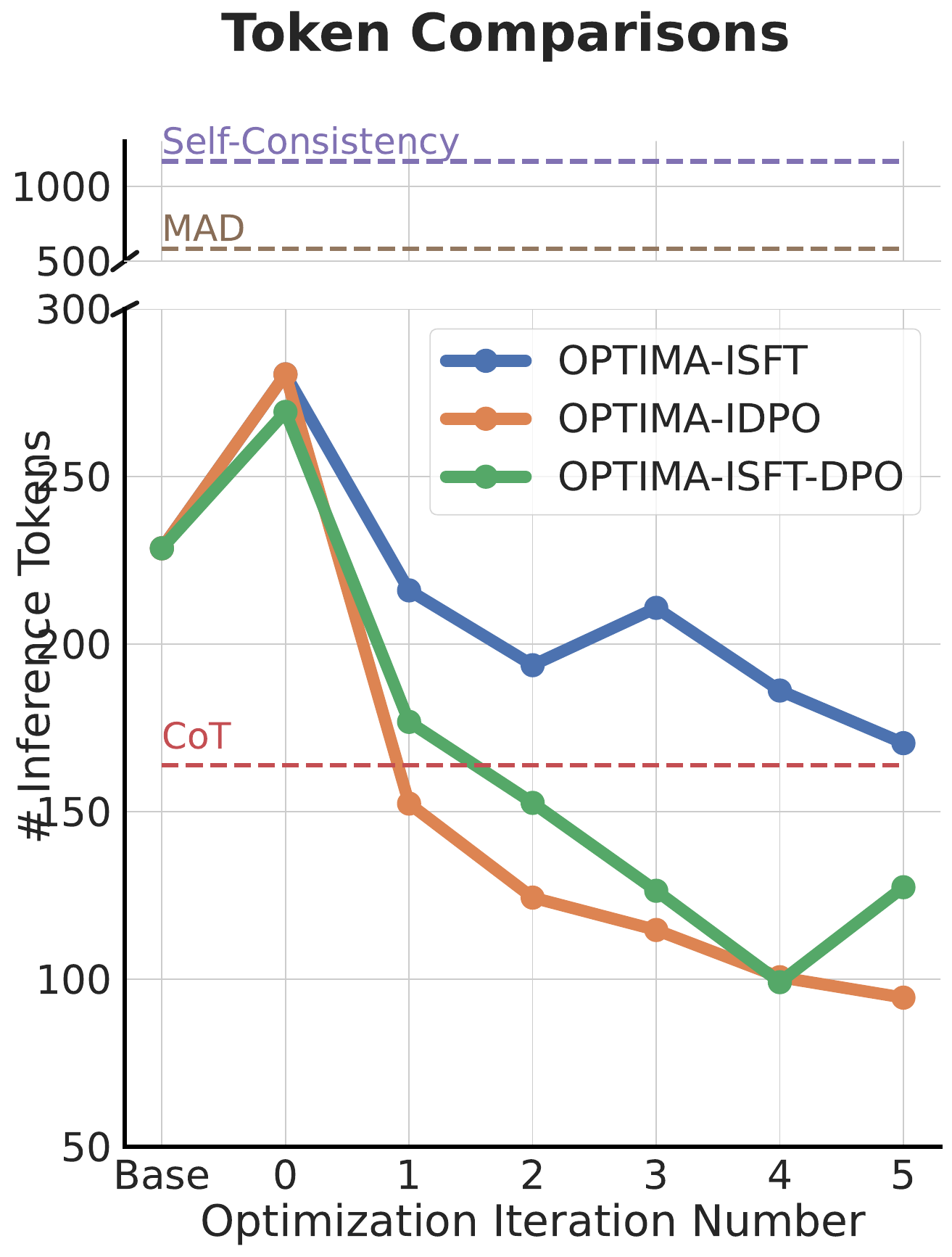}
        % \caption{Caption}
        \label{fig:performance-overview-token}
    \end{subfigure}
    \vspace{-1.5em}
    \caption{\textbf{Performance and efficiency of \optima variants across optimization iterations.} \textbf{Left}: Average performance gain over iterations. \optima variants consistently outperform CoT, Multi-Agent Debate (MAD), and Self-Consistency. \textbf{Right}: Average inference token numbers over iterations. All \optima variants achieve better performance with substantially fewer tokens.}
    \vspace{-1.5em}
    \label{fig:performance-overview}
\end{figure}

\section{Introduction}
\label{sec:introduction}

Large Language Models (LLMs) have emerged as powerful tools for a wide range of tasks, from natural language processing to complex reasoning \citep{DBLP:journals/corr/abs-2303-08774,DBLP:journals/corr/abs-2403-05530,anthropic2024claude}. A promising direction in leveraging these models is the development of autonomous multi-agent systems (MAS), which aim to harness the collective intelligence of multiple LLM-based agents for collaborative problem-solving and decision-making \citep{DBLP:journals/corr/abs-2305-19118,DBLP:conf/naacl/WangMW0WJ24,DBLP:conf/icml/Du00TM24,DBLP:conf/icml/ZhugeWKFKS24}. However, for LLM-based MAS to be truly effective, they must overcome two critical challenges: \textbf{(a)} achieving efficient inter-agent communication to minimize computational costs, and \textbf{(b)} optimizing the collective performance of the system as a cohesive unit.

Current LLM-based MAS face significant difficulties in meeting these challenges. The coordination and communication between agents often lack efficiency, resulting in verbose exchanges that lead to increased token usage, longer inference times, and higher computational costs \citep{DBLP:journals/corr/abs-2406-11776}. This inefficiency is exacerbated by the \textit{length bias} inherent in LLMs due to alignment training \citep{DBLP:journals/corr/abs-2310-10076,DBLP:journals/corr/abs-2404-04475}, which favors longer responses even when concise communication would suffice \citep{DBLP:journals/corr/abs-2402-18439}. Moreover, while recent work has explored training LLMs for single-agent tasks \citep{DBLP:journals/corr/abs-2403-02502,DBLP:journals/corr/abs-2406-11176} and MAS training is well-studied in reinforcement learning \citep{DBLP:journals/ijon/JohnsonLC00,DBLP:conf/nips/LanctotZGLTPSG17,DBLP:conf/iclr/BakerKMWPMM20}, there remains a lack of parameter-updating methods specifically designed to optimize LLM-based MAS as a unified system. Existing approaches primarily rely on simple agent profile evolution \citep{DBLP:conf/iclr/ChenSZ0YCYLHQQC24} or memory evolution \citep{DBLP:conf/acl/QianDLLXWC0CCL024,DBLP:journals/corr/abs-2405-04219,DBLP:journals/corr/abs-2404-05569}, which fail to address the core issues of communication efficiency and collective optimization.

\textbf{Can we develop a training framework that simultaneously enhances the communication efficiency and task effectiveness of LLM-based MAS?} To address this question, we introduce \textbf{\optima}, an effective framework designed to optimize LLM-based MAS. At the heart of \optima is an iterative \textit{generate, rank, select, and train} paradigm, incorporating a reward function that balances task performance, token efficiency, and communication readability. This approach enables the development of MAS that are not only effective and efficient but also maintain interpretable communication patterns. Based on the reward function, \optima leverages a combination of techniques to induce efficient and effective communication behaviors in LLM-based agents, including Supervised Fine-Tuning (SFT) \citep{DBLP:conf/nips/ZelikmanWMG22,DBLP:journals/corr/abs-2308-08998,DBLP:journals/corr/abs-2312-10003} and Direct Preference Optimization (DPO) \citep{DBLP:conf/nips/RafailovSMMEF23,DBLP:journals/corr/abs-2404-19733}, along with their hybrid variants. Furthermore, \optima introduces an integration of Monte Carlo Tree Search (MCTS)-inspired techniques for DPO data generation, conceptualizing conversation turns as tree nodes to explore diverse interaction trajectories efficiently.

Importantly, by substantially reducing the number of tokens required for inference, \optima not only improves computational efficiency but also opens new possibilities for leveraging inference compute more effectively. This reduction in token usage allows for more samples within the same computational constraints, potentially leading to \textit{better inference-time scaling laws}. As recent work has shown the importance of inference-time compute in improving model performance \citep{DBLP:journals/corr/abs-2408-00724,DBLP:journals/corr/abs-2407-21787,DBLP:journals/corr/abs-2403-02419}, \optima's efficiency gains could be combined with techniques like majority voting \citep{DBLP:conf/iclr/0002WSLCNCZ23}, leading to more effective LLM systems.

We evaluate \optima on a diverse set of tasks spanning two multi-agent settings: \textbf{(a)} information exchange, including information-asymmetric question answering \citep{DBLP:journals/corr/abs-2402-18439,DBLP:journals/corr/abs-2406-14928}, and \textbf{(b)} debate, encompassing mathematical and reasoning tasks \citep{DBLP:conf/icml/Du00TM24,DBLP:conf/iclr/ChenSZ0YCYLHQQC24,DBLP:journals/corr/abs-2308-08155}. Using Llama 3 8B / 3.2 3B \citep{llama3modelcard} as our base model, we demonstrate that \optima consistently outperforms both single-agent MAS baselines, achieving up to 90\% reduction in token usage and 2.8x increase in task performance.

To summarize, our main contribution is \optima, a novel training framework that simultaneously optimizes \textit{communication efficiency} and \textit{task effectiveness}. To enhance high-quality training data generation \textit{in multi-agent settings} for DPO, we introduce an integration of MCTS-like techniques. Our comprehensive empirical evaluation across diverse tasks demonstrates notable advancements in \textit{both} token efficiency and task performance, while also providing insights into the learned communication patterns. Additionally, we examine the implications of \optima's efficiency gains for inference-time scaling, underscoring its potential to improve the LLM systems by enabling more effective utilization of inference-compute. By addressing the dual challenges of communication efficiency and collective optimization, our work underscores the importance of developing advanced training frameworks for LLM-based MAS and highlights efficiency as a crucial metric to consider. We believe \optima provides a solid foundation for future investigations into scaling and improving MAS and general LLM systems.

\section{\optima: Optimizing Multi-Agent LLMs via Iterative Training}
\label{sec:method}

\subsection{Overview}
\label{subsec:method-overview}

\begin{figure*}[t!]
    \centering
    \includegraphics[width=0.9\linewidth]{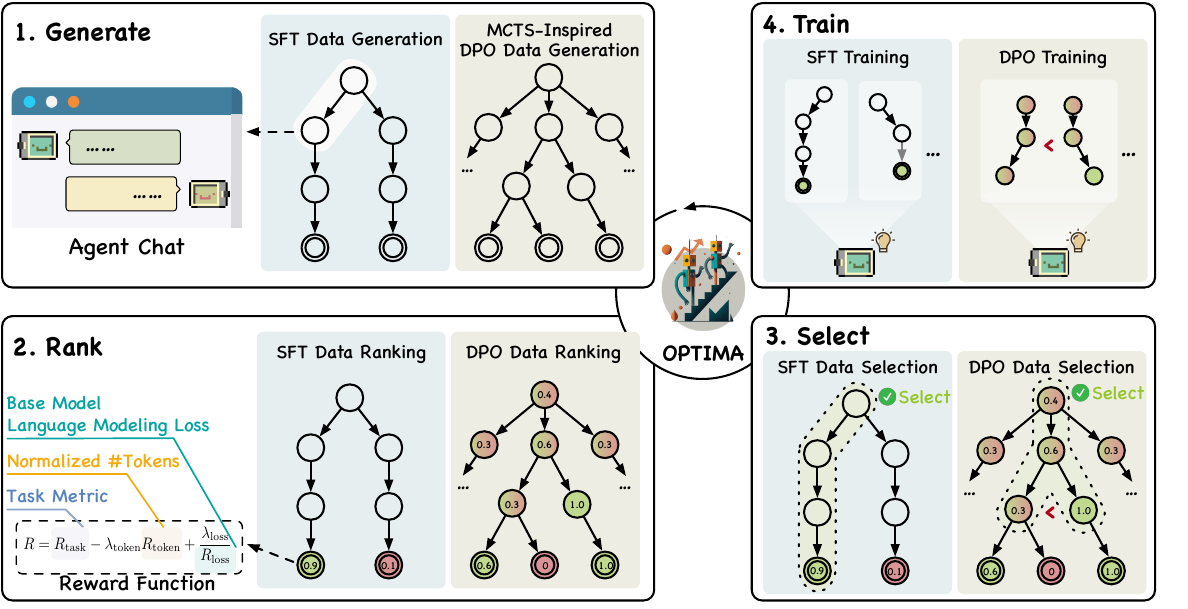}
    % \vspace{-0.8em}
    \caption{\textbf{Overview of the \optima framework for training LLM-based MAS}. The iterative process includes four stages: \textit{Generate, Rank, Select}, and \textit{Train}. Note that the ranking process, while also involved in DPO data generation, is not shown in the Generate stage for simplicity.}
    \label{fig:optima}
    \vspace{-1.3em}
\end{figure*}

\optima is built upon an iterative \textit{generate, rank, select, and train} paradigm. This approach allows for the progressive improvement of LLM-based agents in multi-agent settings, focusing on enhancing both the efficiency of inter-agent communication and the effectiveness of task completion. 

Let $\mathcal{M}_\text{base}$ denote the base LLM, $\mathcal{D}$ the task dataset, and $f$ the iterative training function. The iterative process can be formalized as $\mathcal{M}_{t+1} = f(\mathcal{M}_t, \mathcal{D})$,
where $\mathcal{M}_t$ represents the model at iteration $t$. The function $f$ encapsulates the entire process of data generation, ranking, selection and model training.
For each task instance $d_i \in \mathcal{D}$, we sample a set of $N$ conversation trajectories $\{\tau_i^j\}_{j=1}^N\subset \mathcal{T}$ using the agents powered by current model $\mathcal{M}_t$. Each trajectory $\tau_i^j$ is then evaluated using a reward function $R: \mathcal{T} \rightarrow \mathbb{R}$, defined as:
\begin{equation}
\resizebox{0.89\linewidth}{!}{$R(\tau_i^j) = R_\text{task}(\tau_i^j) - \lambda_\text{token} R_\text{token}(\tau_i^j) + \lambda_\text{loss} \frac{1}{R_\text{loss}(\tau_i^j)}$.}
    \label{eq:reward}
\end{equation}
Here, $R_\text{task}: \mathcal{T} \rightarrow \mathbb{R}$ is the task-specific performance metric, $R_\text{token}(\tau_i^j) = \frac{\#\text{Tokens}(\tau_i^j)}{\max_k(\{\#\text{Tokens}(\tau_i^k)\}_k)}$ is the normalized token count, and $R_\text{loss}(\tau_i^j) = g\big(\mathcal{L}(\mathcal{M}_\text{base}, d_i, \tau_i^j)\big)$ is based on the language modeling loss of the base model $\mathcal{M}_\text{base}$, which we detail in \cref{appendix:details-ranking}. The positive coefficients $\lambda_\text{token}$ and $\lambda_\text{loss}$ are hyper-parameters . This reward function is designed to balance multiple objectives simultaneously: $R_\text{task}$ ensures that the model improves on the intended task, $R_\text{token}$ encourages communication efficiency by penalizing verbose exchanges, and $R_\text{loss}$ regularizes language naturalness and readability by favoring trajectories that are probable under the base model. By incorporating these components, we aim to develop LLM-based MAS that are not only effective in their designated tasks but also efficient in their communication, while maintaining interpretability in their outputs, unlike the often incomprehensible communication in prior RL research \citep{DBLP:conf/iclr/LazaridouPB17,DBLP:conf/iclr/EvtimovaDKC18,DBLP:conf/iclr/ChaabouniSATTDM22}.

Based on these rewards, we apply several data selection criteria to select a subset of high-quality sampled trajectories $\{\tau_i^*\}$ for each task instance. These selected trajectories form the training data $\mathcal{D}_i^*$ at iteration $i$. The model is then updated:
% \begin{equation}
$\mathcal{M}_{t+1} = \text{Train}(\mathcal{M}_t, \mathcal{D}_i^*).$
% \end{equation}
The Train function can be instantiated with various training algorithms, such as SFT or DPO, which we will discuss in detail in the following subsections.

\cref{fig:optima} provides a high-level overview of \optima. The specific instantiations of the generation and training processes will be detailed in the following subsections. The ranking process, consistent across all instantiations, is defined by the reward function presented in \cref{eq:reward}.

\subsection{Initialization}
\label{subsec:method-initialization}

Before starting the iterative training process, we address a critical challenge in LLM-based MAS: agents often produce responses in a similar style across conversation trajectories, even with high-temperature sampling. This homogeneity limits the exploration of diverse communication strategies, potentially hindering the optimization toward more efficient and effective interactions. Following the observation from AutoForm \citep{DBLP:journals/corr/abs-2402-18439}, where LLMs can be explicitly prompted to leverage different more concise formats to communicate or reason without much compromise in performance, we introduce an initialization step that promotes diversity in agent communication.

Our approach leverages a pool of format specification prompts, $\mathcal{P} = \{p_1, p_2, ..., p_K\}$, where each $p_k$ is a string specifying a particular response format (e.g., JSON, list, see \cref{appendix:prompts} for concrete examples and creation process). For each task instance $d_i \in \mathcal{D}$, we generate $N$ conversation trajectories, each with a randomly selected format specification appended to the input task:
\begin{equation}
    \tau_i^j = \mathcal{M}_\text{base}(d_i \oplus p_{k_j}), \ k_j \sim \text{Uniform}(1, K),
\end{equation}
where $\oplus$ denotes string concatenation. This process yields a diverse set of trajectories $\{\tau_i^j\}_{j=1}^N$ for each $d_i$, varying in both content and structure.

We then evaluate these trajectories using the reward function defined in \cref{eq:reward}, for each $d_i$, we select the trajectory with the highest reward: $\tau_i^* = \argmax_j R(\tau_i^j)$. Finally, we select top 70\% trajectories that exceed a predefined performance threshold $\theta_\text{init}$, resulting in a high-quality dataset:
\begin{equation}
\resizebox{0.89\linewidth}{!}{$\mathcal{D}_0^* = \text{TopK} ( \big\{ (d_i, \tau_i^*) \, \big| \, R_\text{task}(\tau_i^*) > \theta_\text{init}, \forall d_i \in \mathcal{D} \big\}, 70\% ).$}
\end{equation}

Crucially, we remove the format specification prompts from the selected trajectories, resulting in a dataset of diverse, high-quality conversations without explicit format instructions. We then fine-tune the base model $\mathcal{M}_\text{base}$ to obtain $\mathcal{M}_0 = \text{SFT}(\mathcal{M}_\text{base}, \mathcal{D}_0^*)$,
which serves as the starting point for \optima, able to generate diverse communication patterns without explicit format prompting.
We provide pseudo-code in \cref{appendix:pseudo-codes} for better understanding.
This initialization sets the stage for more effective exploration and optimization in the subsequent iterative training process.

\subsection{Instantiation 1: Iterative SFT}
\label{subsec:method-iterative-sft}

We introduce iterative Supervised Fine-Tuning (iSFT) as our first instantiation of \optima. At each iteration $t$, iSFT follows the same general procedure outlined in \cref{alg:initialization}, generating a set of $N$ conversation trajectories for each task training instance $d_i \in \mathcal{D}$ using the current model $\mathcal{M}_t^\text{iSFT}$. However, unlike initialization, iSFT omits the format specification pool, as $\mathcal{M}_0$ has already internalized diverse communication strategies. Unlike recent research on iterative training \citep{DBLP:journals/corr/abs-2308-08998,DBLP:journals/corr/abs-2312-10003}, iSFT maintains a fixed reward threshold $\theta_\text{SFT}$ across iterations for data selection. The model is then trained with standard SFT. This process continues until a maximum number of iterations is reached. For clarity, the pseudo-code for iSFT is provided in \cref{appendix:pseudo-codes}.

iSFT provides a straightforward yet effective approach to optimize LLM-based MAS, leveraging the diverse communication patterns established during initialization while consistently improving task performance and communication efficiency.

\subsection{Instantiation 2: Iterative DPO}
\label{subsec:method-iterative-dpo}

While iSFT provides a straightforward approach to optimizing LLM-based MAS, it may be limited by its reliance on a single \textit{best} trajectory for each task instance. To address this, we explore iterative Direct Preference Optimization (iDPO) \citep{DBLP:conf/nips/RafailovSMMEF23,DBLP:journals/corr/abs-2404-19733}, which optimizes models using comparative preferences and has demonstrated success in LLM alignment. Applying DPO in multi-agent settings, however, poses distinct challenges, particularly in generating meaningful paired data that capture the complexities of agent interactions.

\textbf{Data Generation}: To overcome these challenges, we integrate MCTS with DPO data collection for high-quality paired data generation in multi-agent settings. Our MCTS-based approach conceptualizes the multi-agent conversation as a tree, where nodes represent conversational turns, and edges represent continuations. This structure allows us to explore diverse interaction trajectories systematically and select high-quality paired data for DPO training. The MCTS process begins at the root node (initial task prompt) and proceeds as follows:
\textbf{(1) Expansion}: We select a node to expand based on the following criteria. We first exclude leaf nodes and the second-to-last level nodes to avoid wasting computation on low-variance expansions, then exclude nodes with content similar to previously expanded nodes, measured based on edit distance (see \cref{appendix:details-data-generation}). From the remaining nodes, we select 10 nodes with the highest rewards and sample one using the softmax distribution over their rewards.
\textbf{(2) Simulation}: For each selected node, we expand 3 trajectories, simulating the conversation to completion.
\textbf{(3) Backpropagation}: Once a trajectory is completed and rewarded with \cref{eq:reward}, we update the estimated rewards of all nodes in the trajectory with the average rewards from their children.
\textbf{(4) Iteration}: We repeat the above process 8 times, resulting in 24 trajectories. More iterations could potentially lead to more diverse and better-quality data.

\textbf{Paired Data Construction}: To generate high-quality paired data for DPO training, we traverse each MCTS tree and identify node pairs $(n_i, n_j)$ that satisfy three conditions: (1) shared ancestry, (2) the higher estimated reward of $n_i$ and $n_j$ exceeds the threshold $\theta_\text{dpo-filter}$, and (3) their reward difference exceeds the threshold $\theta_\text{dpo-diff}$. We sort these pairs by the higher estimated reward, and select the top 50\% pairs as part of the final training set. We construct DPO training instances by using the common conversation history as the prompt, with $n_i$ and $n_j$ serving as the chosen and rejected responses according to their estimated rewards. 

The iDPO process then proceeds iteratively, alternating between MCTS-based data generation and model updates using DPO. The pseudo-code for our iDPO process is presented in \cref{appendix:pseudo-codes}.

\subsection{Instantiation 3: Hybrid Iterative Training}
\label{subsec:method-hybrid}

Building upon the strengths of both iSFT and iDPO, we investigate a hybrid approach that interleaves SFT and DPO in the iterative training process, termed as iSFT-DPO. This hybrid method aims to leverage the simplicity and directness of SFT in capturing high-quality trajectories, while also benefiting from the nuanced comparative learning facilitated by DPO. By alternating between these two training paradigms, we hypothesize that the model can more effectively balance the exploration of diverse communication strategies with the exploitation of known effective patterns.

In practice, we implement this hybrid approach by performing one iteration of iSFT followed by one iteration of iDPO, and repeating this cycle throughout the training process. This interleaving allows the model to first consolidate learning from the best observed trajectories through SFT, and then refine its understanding through the comparative preferences provided by DPO.
\section{Experiments}
\label{sec:experiments}

\textbf{Datasets.} We evaluate \optima in two settings: information exchange (IE) and debate. For IE, we use HotpotQA \citep{DBLP:conf/emnlp/Yang0ZBCSM18}, 2WikiMultiHopQA (2WMHQA) \citep{DBLP:conf/coling/HoNSA20}, TriviaQA \citep{DBLP:conf/acl/JoshiCWZ17}, and CBT \citep{DBLP:journals/corr/HillBCW15}. For multi-hop datasets (HotpotQA, 2WMHQA), we split relevant contexts between two agents, ensuring the answer can only be deduced from information exchange. For TriviaQA and CBT, contexts are randomly assigned, challenging agents to communicate and identify the relevant information. The debate setting employs GSM8K \citep{DBLP:journals/corr/abs-2110-14168}, MATH \citep{DBLP:conf/nips/HendrycksBKABTS21}, ARC's challenge set (ARC-C) \citep{DBLP:journals/corr/abs-2102-03315} and MMLU \citep{DBLP:conf/iclr/HendrycksBBZMSS21}, with one agent as solver and another as critic \citep{DBLP:conf/iclr/ChenSZ0YCYLHQQC24}. We use 0-shot for all benchmarks.

\textbf{Metrics.} We report F1 score between generated answers and labels for IE tasks. For debate tasks, we employ exact match accuracy (GSM8k, ARC-C, MMLU) or Sympy-based \citep{DBLP:journals/peerj-cs/MeurerSPCKRKIMS17} equivalence checking (MATH), following \citet{DBLP:conf/nips/LewkowyczADDMRS22}. Conversations conclude when agents both mark the same answer with specified special tokens or reach a turn limit.

\textbf{Baselines.} We compare against single-agent approaches: Chain-of-Thought (CoT) \citep{DBLP:conf/nips/Wei0SBIXCLZ22} and Self-Consistency (SC) with majority voting \citep{DBLP:conf/iclr/0002WSLCNCZ23} on $n=8$ samples. For IE tasks, direct majority voting is impractical due to free-form responses. Instead, we compute pairwise F1 scores, group answers with scores above 0.9, and report the average F1 score of the largest group against the label. In multi-agent settings, we compare against Multi-Agent Debate (MAD) \citep{DBLP:conf/icml/Du00TM24} and AutoForm \citep{DBLP:journals/corr/abs-2402-18439}. MAD uses natural language for inter-agent communication, while AutoForm employs concise, non-natural-language formats for better performance-cost efficiency.

\textbf{Training Setups.} We use Llama 3 8B / 3.2 3B \citep{llama3modelcard} as our base model, focusing on two-agent scenarios without external tools to isolate core multi-agent communication and collaboration. A single model is trained for both agents, with separate model training left for future work. Iterative training completes within 12 hours on 8 A100 GPUs for most tasks, except MATH, which takes around 24 hours. More details are in \cref{appendix:exp-detail,appendix:prompts}.

\subsection{Benchmark Results}
\label{subsec:experiments-results}
\begin{table*}[t]
    \centering
    \vspace{-0.8em}
    \resizebox{\linewidth}{!}{
    \setlength{\tabcolsep}{3pt}
    \renewcommand{\arraystretch}{1.1}
    \begin{tabular}{l*{8}{cr}}
    \toprule
    & \multicolumn{8}{c}{\textbf{Information Exchange}} & \multicolumn{8}{c}{\textbf{Debate}} \\
    \cmidrule(lr){2-9} \cmidrule(lr){10-17}
     & \multicolumn{2}{c}{\textbf{HotpotQA}} & \multicolumn{2}{c}{\textbf{2WMH QA}} & \multicolumn{2}{c}{\textbf{TriviaQA}} & \multicolumn{2}{c}{\textbf{CBT}} & \multicolumn{2}{c}{\textbf{MATH}} & \multicolumn{2}{c}{\textbf{GSM8k}} & \multicolumn{2}{c}{\textbf{ARC-C}}&\multicolumn{2}{c}{\textbf{MMLU}} \\
    \cmidrule(lr){2-3} \cmidrule(lr){4-5} \cmidrule(lr){6-7} \cmidrule(lr){8-9} \cmidrule(lr){10-11} \cmidrule(lr){12-13} \cmidrule(lr){14-15} \cmidrule(lr){16-17}   
    \textbf{Method} & \multicolumn{1}{c}{\textbf{F1}} & \multicolumn{1}{c}{\textbf{\#Tok}} & \multicolumn{1}{c}{\textbf{F1}} & \multicolumn{1}{c}{\textbf{\#Tok}} & \multicolumn{1}{c}{\textbf{F1}} & \multicolumn{1}{c}{\textbf{\#Tok}} & \multicolumn{1}{c}{\textbf{F1}} & \multicolumn{1}{c}{\textbf{\#Tok}} & \multicolumn{1}{c}{\textbf{Acc}} & \multicolumn{1}{c}{\textbf{\#Tok}} & \multicolumn{1}{c}{\textbf{Acc}} & \multicolumn{1}{c}{\textbf{\#Tok}} & \multicolumn{1}{c}{\textbf{Acc}} & \multicolumn{1}{c}{\textbf{\#Tok}} & \multicolumn{1}{c}{\textbf{Acc}} & \multicolumn{1}{c}{\textbf{\#Tok}} \\
    % single
    \midrule
    CoT & 25.6 &123.7 &20.5 &139.8 &59.8 &110.3 &43.4&135.3& 23.9 & 329.8 & 71.5 &230.9 & 65.2 &138.9 & 46.0 & 132.2\\
    SC ($n=8$) & 33.8 & 996.3 &28.7 &1052.8 &70.0 &891.4&52.9 &1067.7 & \textbf{35.7} & 2600.9 & \underline{80.3} & 1828.7 & \underline{75.6} & 1116.7 & 54.0 & 1056.1\\
    % multi
    \midrule
    MAD &  28.4 & 570.9 &25.9&543.7 & 71.0 & 408.6 & 53.8 & 493.0& 29.8 & 1517.6 & 72.5 & 514.7 & 71.4 &478.0 & 51.5 & 516.7\\
    AutoForm & 28.2 & 97.7 &24.7 & 117.7 & 60.9 & 74.0 & 35.0 &64.8& 26.1 & 644.3 & 71.0 &410.5 & 60.2 & 221.2 & 43.8 & 198.5\\
    % train
    \midrule
    \optima-iSFT & \underline{54.5} & 67.6  & \underline{72.4} & 61.2 & \underline{71.9} & \underline{51.5} & \textbf{71.8} & \underline{38.5}& 30.1 & 830.3 & 79.5 & 311.5& 74.1 & \underline{92.2} & 56.8 & 123.8\\
    \optima-iDPO & 52.5 & \textbf{45.7} &66.1&\textbf{35.9} &69.3 & 69.2 & 66.7 & \textbf{37.2}& \underline{30.4} & \textbf{272.8} & 78.5 & \underline{270.1} & 74.5 & 97.8 & \underline{59.6} & \underline{61.6}\\
    \optima-iSFT-DPO &  \textbf{55.6} & \underline{63.3} & \textbf{74.2} & \underline{54.9} & \textbf{77.1} & \textbf{32.5} & \underline{70.1} & 38.9& 29.3 & \underline{488.1} & \textbf{80.4} & \textbf{246.5} & \textbf{77.1} & \textbf{88.0} & \textbf{60.2} & \textbf{56.7}\\
    \midrule
    \optima-iSFT SC & 54.8 & 806.2 & 72.6 & 245.6 & 73.7& 413.8 & \colorbox{mygreen}{72.2} & 847.4 &32.4 & 2432.9 & 83.1 & 1750.7 & 77.2 &1148.7 & 60.2 &874.5\\
    \optima-iDPO SC &  52.8 & 412.8& 67.2& 1056.2& 71.8& 702.8& 66.8& 520.6& \colorbox{mygreen}{36.9} & 2743.1 & \colorbox{mygreen}{84.4} &1750.8 & 77.0 & 1091.2 & 59.9 & 1050.4\\
    \optima-iSFT-DPO SC & \colorbox{mygreen}{57.4} & 957.9 & \colorbox{mygreen}{76.7} &1096.0 & \colorbox{mygreen}{77.5} & 494.1 & 71.8 & 417.8 & 34.8 & 2788.5 & 84.0 & 1748.7& \colorbox{mygreen}{78.8} & 1036.1 & \colorbox{mygreen}{61.2} & 1026.7 \\
    \bottomrule
    \end{tabular}}
    \vspace{-0.7em}
    \caption{\textbf{Performance and inference token number comparison across information exchange and debate tasks.} Best results are indicated in \textbf{bold}, and second-best results are \underline{underlined} for all rows except the last three. The last three rows display self-consistency results for \optima variants, with the best results highlighted in\colorbox{mygreen}{green}. \optima variants consistently outperform baselines in task performance and/or token efficiency.}
    \vspace{-1.5em}
    \label{tab:main-table}
    % \vspace{-1.5em}
\end{table*}

\begin{table}
    \centering
    \resizebox{\linewidth}{!}{
    \begin{tabular}{l*{6}{r}}
    \toprule
     & \multicolumn{2}{c}{\textbf{2WMH QA}} & \multicolumn{2}{c}{\textbf{Trivia QA}} & \multicolumn{2}{c}{\textbf{GSM8k}} \\
    \cmidrule(lr){2-3} \cmidrule(lr){4-5} \cmidrule(lr){6-7}
    \textbf{Method} & \multicolumn{1}{c}{\textbf{F1}} & \multicolumn{1}{c}{\textbf{\#Tok}} & \multicolumn{1}{c}{\textbf{F1}} & \multicolumn{1}{c}{\textbf{\#Tok}} & \multicolumn{1}{c}{\textbf{Acc}} & \multicolumn{1}{c}{\textbf{\#Tok}}\\
    % multi
    \midrule
    MAD & 25.9 & 543.7 & 71.0 &408.9 & 72.5 & 514.7\\
    AutoForm & 24.7 & 117.7 & 60.9 & 74.0 & 71.0 & 410.5 \\
    % train
    \midrule
    iSFT & \textbf{56.5} & 79.6 &  70.0 & 90.2 & 74.6 & 293.7 \\
    iDPO & 51.6 & 84.3 & 68.0 & \textbf{41.1} & \textbf{77.9} & \textbf{185.7}\\
    iSFT-DPO & 54.5 & \textbf{70.4} & \textbf{72.0} & 67.8 & 74.2 & 363.1 \\
    \bottomrule
    \end{tabular}
    }
    \vspace{-0.7em}
    \caption{\textbf{Transfer performance of \optima.} We transfer \optima from Hotpot QA to 2WMH QA and Trivia QA, and from MATH to GSM8k, with MAD and AutoForm on each target task as baselines.}
    \vspace{-1.5em}
    \label{tab:transfer}
% \end{wraptable}
% \end{table*}
\end{table}

\cref{tab:main-table} showcases \optima's performance across diverse tasks, revealing consistent improvements in effectiveness and efficiency. For IE tasks, \optima variants excel, particularly in multi-hop reasoning like HotpotQA and 2WMHQA. iSFT-DPO achieves the best performance while significantly reducing token usage compared to SC. Notably, on 2WMHQA, iSFT-DPO improves F1 by \textbf{38.3\%} (2.8x) while using only \textbf{10\%} of MAD's tokens. This efficiency extends to other IE tasks, where \optima variants maintain high performance with drastically lower token counts.

In debate tasks, \optima's benefits are nuanced but evident. It achieves better performance and efficiency on ARC-C and MMLU, while on MATH and GSM8k, \optima variants show comparable or slightly lower performance than SC, but with much higher token efficiency. We attribute this to task difficulty and limited training data. Nevertheless, \cref{subsec:experiments-transfer} will show \optima models trained on MATH transfer effectively to GSM8k, achieving near-equivalent performance with high efficiency. Additionally, \cref{subsec:experiments-inference-scaling-law} will demonstrate that applying SC to \optima variants trained on MATH or GSM8k greatly improves inference scaling laws on GSM8k compared to CoT SC.

Among \optima variants, iSFT often prioritizes performance at the cost of efficiency, while iDPO achieves remarkable token reductions, sometimes with performance trade-offs. iSFT-DPO strikes a robust balance, frequently delivering top-tier performance with satisfying efficiency. Results on Llama 3.2 3B in \cref{appendix:results-on-llama-3.2-3b} further validate \optima's robustness.

\subsection{How Well Does \optima Generalize?}
\label{subsec:experiments-transfer}

To assess \optima's ability to generalize, we conducted transfer learning experiments across different task domains. We transferred models trained on HotpotQA to TriviaQA and 2WMHQA, as well as transferring from MATH to GSM8k. While these datasets share broad categories (question-answering and mathematical reasoning, respectively), they present different challenges in terms of complexity and required skills. The results, presented in \cref{tab:transfer}, demonstrate \optima's robust transferability across these diverse tasks. In the question-answering domain, all \optima variants significantly outperform baseline multi-agent methods on both OOD datasets. On 2WMHQA, the transferred iSFT more than doubles MAD's F1 score while using only 14.6\% of the tokens. Similar trends are observed in TriviaQA. When transferring from MATH to GSM8k, \optima variants, particular iDPO, not only outperform the baselines on GSM8k but also achieve results comparable to models directly trained on GSM8k with even higher token efficiency (refer to \cref{tab:main-table} for comparison). 

These results underscore \optima's potential for developing adaptable MAS, demonstrating that \optima-trained models learn transferable skills for efficient information exchange and collaborative reasoning. However, transferring to more distant domains remains challenging, e.g., we find it hard to transfer from from MATH to ARC-C. We believe it is a promising area for future research to explore if scaling \optima to more generalized multi-task training could enhance the generalization.

\subsection{Can \optima Improve Inference Scaling?}
\label{subsec:experiments-inference-scaling-law}

\begin{figure}[t]
\centering
\begin{subfigure}{0.48\linewidth}
  \centering
  \includegraphics[width=\linewidth]{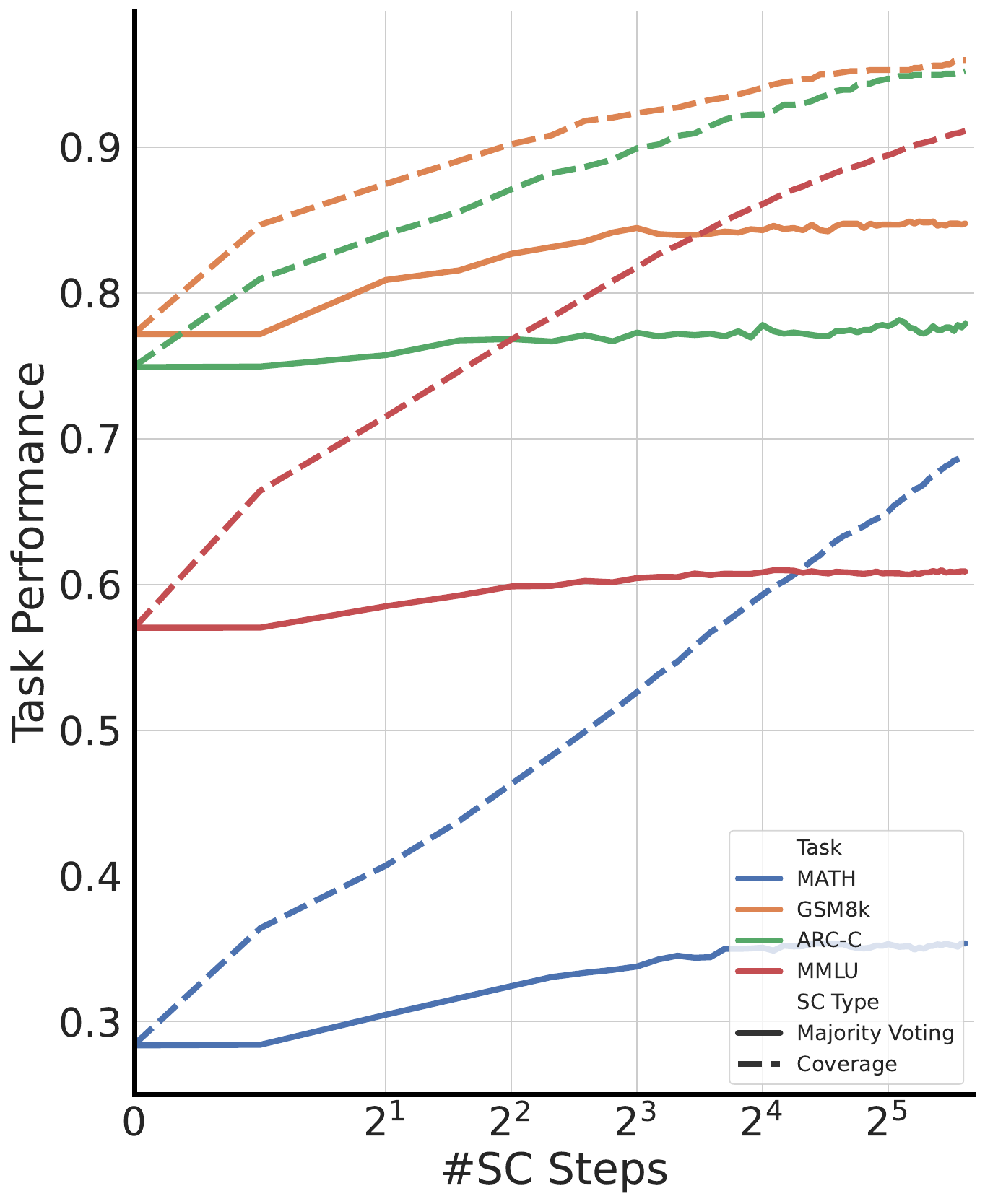}
  % \caption{Inference scaling on debate tasks}
  \vspace{-1.5em}
  \label{fig:debate-inference-scaling-law}
\end{subfigure}%
\begin{subfigure}{0.48\linewidth}
  \centering
  \includegraphics[width=\linewidth]{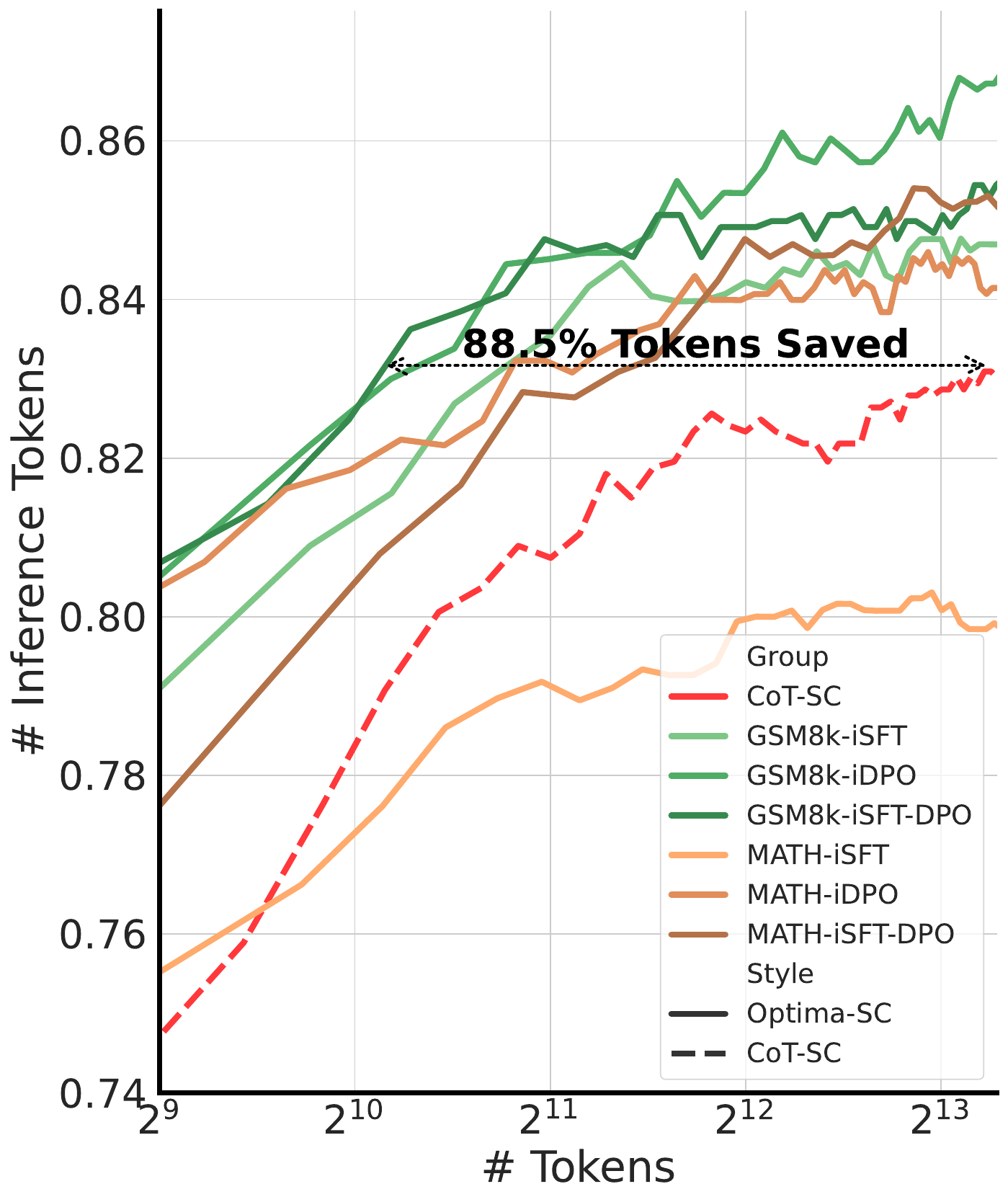}
  % \caption{Performance vs. token usage on GSM8k}
  \vspace{-1.2em}
  \label{fig:gsm8k-inference-scaling-law}
\end{subfigure}
\vspace{-0.6em}
\caption{\textbf{\optima's impact on inference scaling laws.} \textbf{Left} Relationship between \optima variants' self-consistency steps and performance on debate tasks. \underline{Solid lines} represent majority voting accuracy, while \dashuline{dashed lines} show coverage. \textbf{Right} Performance of various models on GSM8k as a function of token usage, demonstrating \optima's efficiency gains.}
\label{fig:inference-scaling}
\vspace{-1.5em}
\end{figure}

Recent research emphasizes inference-time scaling, which describes how model performance improves with increased compute during inference, typically by generating multiple samples per problem \citep{DBLP:journals/corr/abs-2407-21787,DBLP:journals/corr/abs-2408-00724}. Unlike training scaling laws, which focus on model size, dataset size, and performance, inference-time scaling explore the trade-off between compute budget and task accuracy, offering a promising way to enhance model capabilities without further training.

\cref{fig:inference-scaling} illustrates \optima's impact on inference-time scaling. The left panel shows the relationship between SC steps and performance on multi-agent debate tasks. While majority voting accuracy plateaus after a certain number of steps, coverage (the percentage of problems answered correctly at least once) improves logarithmically with increased sampling. This aligns with recent studies \citep{DBLP:journals/corr/abs-2408-00724,DBLP:journals/corr/abs-2403-02419}, suggesting advanced answer selection techniques could further boost \optima's performance. Additional scaling law figures for all \optima variants and tasks are in \cref{appendix:info-exchange-scaling}, where similar trends are observed.

The right panel demonstrates \optima's efficiency in improving inference scaling laws on GSM8k. \optima variants, including those transferred from MATH, consistently outperform CoT SC, except for MATH-trained iSFT. Notably, GSM8k-trained iDPO matches CoT-SC performance with 88.5\% fewer tokens, effectively ``\textit{shifting the curve left}". This reduction in token usage translates to significant computational savings without sacrificing accuracy. MATH-trained \optima variants, except iSFT, also deliver better scaling laws on GSM8k than CoT SC, highlighting \optima's cross-task generalization.

These results underscore \optima's potential to reshape inference-time scaling for MAS and general LLM systems. By enabling more efficient use of compute budgets, \optima achieves better performance at lower costs or higher performance at the same cost. This efficiency opens possibilities for integrating advanced inference techniques like weighted voting or tree-search \citep{DBLP:journals/corr/abs-2408-00724}, potentially leading to further performance gains.

\subsection{How Does \optima Evolve Performance?}
\label{subsec:experiments-ablation}

To understand the impact of reward function components in our reward function, we conducted an ablation study on 2WMHQA (IE) and ARC-C (debate). We removed either token count regularization (\#Tokens) or LM loss (Loss) to address: \textbf{(1)} \textit{How does token count regularization affect efficiency-performance trade-offs?} \textbf{(2)} \textit{What role does LM loss play in maintaining communication quality?} Our findings highlight the importance of each component in balancing performance, efficiency, and language quality.

\cref{tab:ablation-table} presents the results of our ablation study. Removing the token count leads to a substantial increase in the number of generated tokens across settings, with a particularly pronounced effect in the debate task. While this increased verbosity occasionally results in marginal performance improvements, it comes at a significant computational cost. Conversely, eliminating the LM loss results in a decrease in token usage, often producing the most concise outputs among all variants. Examples comparing communication with and without LM loss can be found in \cref{appendix:reward-case}. Without LM loss, the model often generates overly concise messages containing insufficient information and is prone to hallucination, potentially explaining the inferior performance. Overall, \optima's reward function achieves the balance among task effectiveness, token efficiency and dialogue quality, enabling effective and efficient multi-agent collaboration.

% \begin{wraptable}{r}{0.5\linewidth}
\begin{table}[t]
    % \vspace{-1em}
    % \vspace{-1.2em}
    \centering
    \setlength{\tabcolsep}{2pt}
    \resizebox{\linewidth}{!}{
    \begin{tabular}{l*{4}{l}}
    \toprule
     & \multicolumn{2}{c}{\textbf{2WMH QA}} & \multicolumn{2}{c}{\textbf{ARC-C}} \\
    \cmidrule(lr){2-3} \cmidrule(lr){4-5}
    \textbf{Setting} & \textbf{F1} & \textbf{\#Tok} & \textbf{Acc} & \textbf{\#Tok}  \\
    % single
    \midrule
    iSFT & \textbf{72.4} & 61.2 & 74.1 & 92.2\\
    \rowcolor{gray!10}
    \ \ w/o \#Tokens & \textbf{72.4}$_{(\text{0.0})}$ & 290.3$\color{red}_{(\text{4.8x})}$ & \textbf{74.2}$\color{teal}_{(\text{+0.1})}$  & 579.6$\color{red}_{(\text{6.3x})}$ \\
    \rowcolor{gray!10}
    \ \ w/o Loss & 69.7$\color{red}_{(\text{-2.7})}$ & \textbf{45.4}$\color{teal}_{(\text{0.7x})}$ & 72.6$\color{red}_{(\text{-1.5})}$ & \textbf{69.7}$\color{teal}_{(\text{0.8x})}$ \\
    \midrule
    iDPO & 66.1 & \textbf{35.9} & 74.5 & 97.8 \\
    \rowcolor{gray!10}
    \ \ w/o \#Tokens & \textbf{72.9}$\color{teal}_{(\text{+6.8})}$ & 183.3$\color{red}_{(\text{5.1x})}$ & \textbf{75.5}$\color{teal}_{(\text{+1.0})}$ & 266.0$\color{red}_{(\text{2.7x})}$\\
    \rowcolor{gray!10}
    \ \ w/o Loss & 63.0$\color{red}_{(\text{-3.1})}$ &54.6$\color{red}_{(\text{1.5x})}$ & 74.4$\color{red}_{(\text{-0.1})}$ & \textbf{81.2}$\color{teal}_{(\text{0.8x})}$\\
    \midrule
    iSFT-DPO & \textbf{74.2} & 54.9 & \textbf{77.1} & 88.0 \\
    \rowcolor{gray!10}
    \ \ w/o \#Tokens & 63.5$\color{red}_{(\text{-10.7})}$ & 219.7$\color{red}_{(\text{4.0x})}$ & 76.9$\color{red}_{(\text{-0.2})}$ & 354.8$\color{red}_{(\text{4.0x})}$\\
    \rowcolor{gray!10}
    \ \ w/o Loss & 66.7$\color{red}_{(\text{-7.5})}$ & \textbf{38.1}$\color{teal}_{(\text{0.7x})}$ & 76.3$\color{red}_{(\text{-0.8})}$ &  \textbf{63.4}$\color{teal}_{(\text{0.7x})}$\\
    \bottomrule
    \end{tabular}}
    \vspace{-0.6em}
    \caption{Ablation study on reward components for \optima variants on two representative tasks.}
    \vspace{-1.5em}
    \label{tab:ablation-table}
\end{table}
% \end{wraptable}

\subsection{How Agent Communication Evolves over Optimization Iterations?}
\label{subsec:experiments-dynamics}

\cref{fig:performance-overview} shows the performance gains and token efficiency of \optima variants across optimization iterations, revealing a two-phase pattern. In the initial phase (iterations 0-1), all variants show significant performance improvements alongside increased token usage, indicating \optima prioritizes effectiveness by enabling agents to develop sophisticated strategies through expanded communication. In later iterations, \optima refines these strategies for efficiency without sacrificing performance, with token usage decreasing gradually while performance continues to improve.

Concrete examples of \optima's impact on agent communication are provided in \cref{appendix:information-exchange-case} (iSFT on an information exchange task) and \cref{appendix:debate-case} (debate task). The base model tends to produce verbose, repetitive exchanges, while \optima-trained models exhibit more concise and task-oriented communication.

\subsection{Can \optima Scale with More Agents?}
\label{subsec:experiments-more-agents}
While the previous experiments highlight \optima's effectiveness in two-agent scenarios, which is a controlled setting that circumvents issues such as communication order and effectively validates the framework, we also evaluate its scalability in three-agent settings for IE and debate tasks. The results, detailed in \cref{appendix:more-agents}, demonstrate that \optima continues to enhance both effectiveness and efficiency.
\section{Related Work}
\label{sec:related-work}

\textbf{LLM-Based MAS}. LLM-powered multi-agent systems have demonstrated success in collaborative problem-solving through approaches like multi-agent debate \citep{DBLP:journals/corr/abs-2305-19118,DBLP:conf/icml/Du00TM24}. Subsequent work explores role-playing for reasoning \citep{DBLP:conf/naacl/WangMW0WJ24,DBLP:conf/iclr/ChenSZ0YCYLHQQC24}, software development \citep{DBLP:conf/acl/QianLLCDL0CSCXL24,DBLP:conf/iclr/HongZCZCWZWYLZR24}, and embodied interactions \citep{DBLP:conf/iclr/ZhangDSZDTSG24,DBLP:conf/icra/MandiJS24}, with scale and diversity improving performance \citep{DBLP:journals/corr/abs-2406-04692,DBLP:journals/corr/abs-2402-05120}. However, efficiency challenges emerge as systems grow \citep{DBLP:journals/corr/abs-2402-18439,DBLP:journals/corr/abs-2406-07155}, with existing methods focusing on memory updates rather than comprehensive training \citep{DBLP:conf/acl/QianDLLXWC0CCL024}. Our framework addresses this gap through joint optimization of communication efficiency and task effectiveness.

\textbf{Iterative Refinement of LLMs}.
Continual improvement in LLMs has led to various iterative refinement paradigms. Self-reflection mechanisms like Reflexion \citep{DBLP:conf/nips/ShinnCGNY23} and self-refine \citep{DBLP:conf/nips/MadaanTGHGW0DPY23} show promise but are limited by LLMs' self-correction abilities \citep{DBLP:conf/iclr/0009CMZYSZ24,DBLP:conf/iclr/OlaussonIW0S24,DBLP:journals/corr/abs-2406-01297}. More robust approaches, such as ReST \citep{DBLP:journals/corr/abs-2308-08998}, ReST$^\text{EM}$ \citep{DBLP:journals/tmlr/SinghCAAPGLH0XP24}, and STaR \citep{DBLP:conf/nips/ZelikmanWMG22}, fine-tune models on self-generated high-quality reasoning paths. \citet{DBLP:journals/corr/abs-2404-19733} further integrate incorrect paths and train models with DPO. These methods have been extended to complex tasks \citep{DBLP:journals/corr/abs-2312-10003}, but iterative refinement in LLM-based MAS remains underexplored, as does the trade-off between effectiveness and efficiency. Our work addresses this gap by introducing the first effective training framework for iterative optimization in MAS contexts and systematically shedding light on the trade-offs between effectiveness and efficiency.

\textbf{Inference-Time Scaling and Token Efficiency}. 
Compute scaling has enhanced LLM capabilities, with approaches like majority voting and reward-guided tree search improving performance on reasoning tasks \citep{DBLP:journals/corr/abs-2403-02419,DBLP:journals/corr/abs-2408-00724,DBLP:journals/corr/abs-2407-21787,DBLP:journals/corr/abs-2409-15254}. However, these methods increase computational demands, highlighting the need for token efficiency. Recent work achieves efficiency through latent space reasoning via step distillation \citep{DBLP:journals/corr/abs-2311-01460,DBLP:journals/corr/abs-2405-14838,DBLP:journals/corr/abs-2412-06769,DBLP:journals/corr/abs-2412-13171}, but at the cost of comprehensibility. Our framework advances this by (1) demonstrating iterative training framework that improves both token efficiency and task effectiveness in MAS context, and (2) showing that enhanced efficiency enables more sampling within fixed compute budgets, leading to better inference-time scaling.

% \textbf{Efficient Agent System.}

\section{Conclusion}
\label{sec:conclusion}
We introduce \optima, a novel framework for training LLM-based MAS that significantly enhances communication efficiency and task performance. Experiments show \optima's consistent superiority over single-agent and multi-agent baselines. We introduce key innovations such as iterative training, a balanced reward function, and MCTS-inspired data generation. Crucially, \optima effectively improves inference-time scaling and transfers effectively to OOD tasks, underscoring the importance of efficient communication in MAS and LLM systems. While \optima marks a major step forward in multi-agent LLM training, further exploration into its scalability to larger models and more complex scenarios is a promising direction for future research.
\section*{Limitations}
While \optima demonstrates significant improvements in communication efficiency and task effectiveness for LLM-based multi-agent systems, our study has several limitations. \textbf{First}, our experiments primarily focus on two-agent scenarios with a shared model architecture, leaving open questions about scaling to larger teams (e.g., 5-10 agents) and heterogeneous agent configurations. Although preliminary results  with three agents show promising trends (\cref{subsec:experiments-more-agents}), the dynamics of larger groups may introduce new challenges in coordination efficiency that require further investigation. \textbf{Second}, while we demonstrate cross-task generalization within similar domains (e.g., MATH to GSM8k), transferring \optima-trained models to substantially different application areas (e.g., from QA to math or coding) remains unexplored. \textbf{Finally}, while we evaluate on standard benchmarks, real-world deployment scenarios may involve additional constraints that our framework does not explicitly address. These limitations highlight valuable directions for future research rather than fundamental flaws, as \optima's core contributions, iterative optimization with efficiency-aware rewards and MCTS-inspired data generation, provide a flexible foundation adaptable to these extensions.

% Bibliography entries for the entire Anthology, followed by custom entries
%\bibliography{anthology,custom}
% Custom bibliography entries only
\bibliography{custom}

\appendix

\section{Inference Scaling Laws on Information Exchange Tasks}
\label{appendix:info-exchange-scaling}
\begin{figure*}[t]
    \centering
    \begin{subfigure}{0.32\textwidth}
      \centering
      \includegraphics[width=\linewidth]{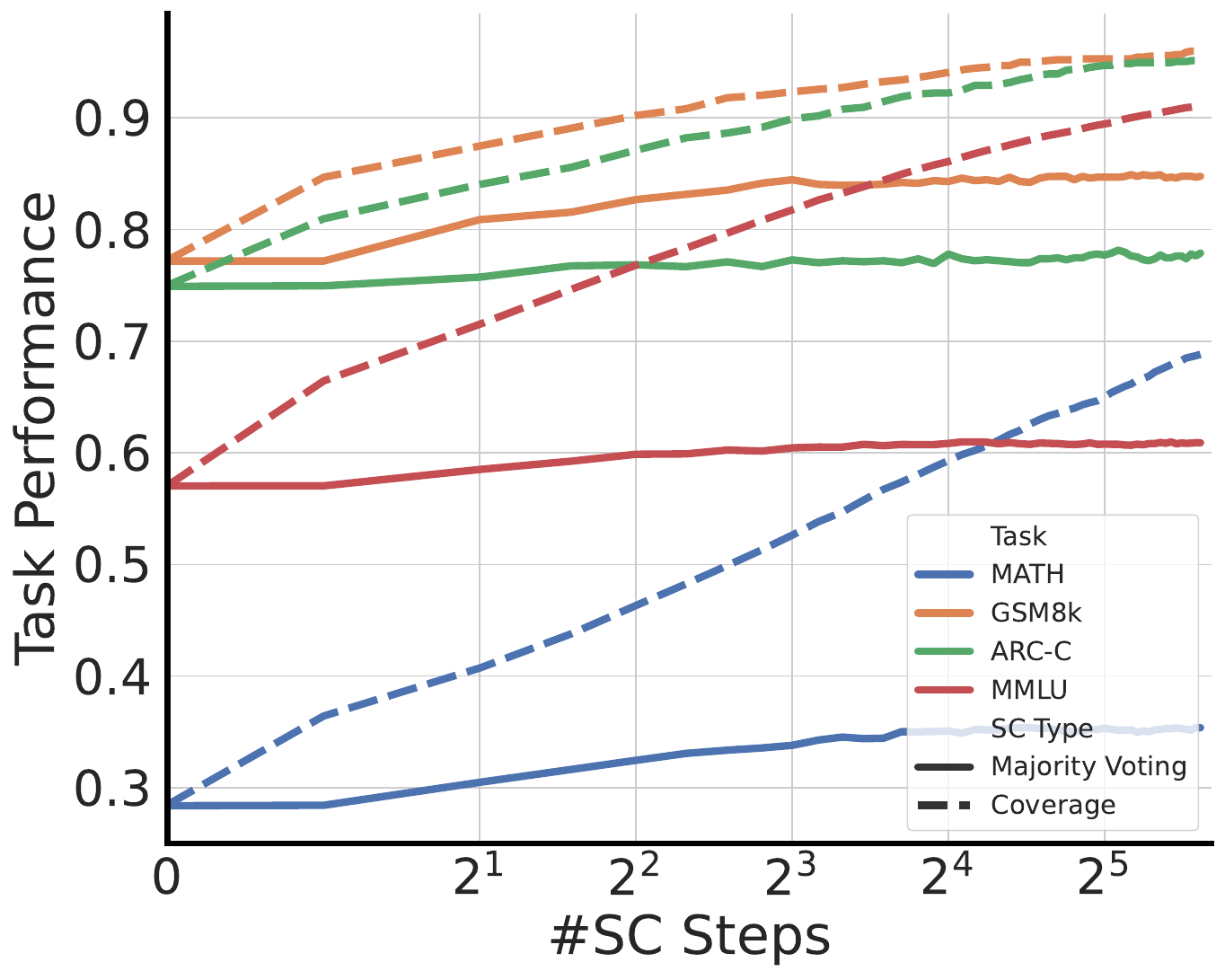}
      \caption{iSFT on Debate tasks.}
    \end{subfigure}
    \begin{subfigure}{0.32\textwidth}
      \centering
      \includegraphics[width=\linewidth]{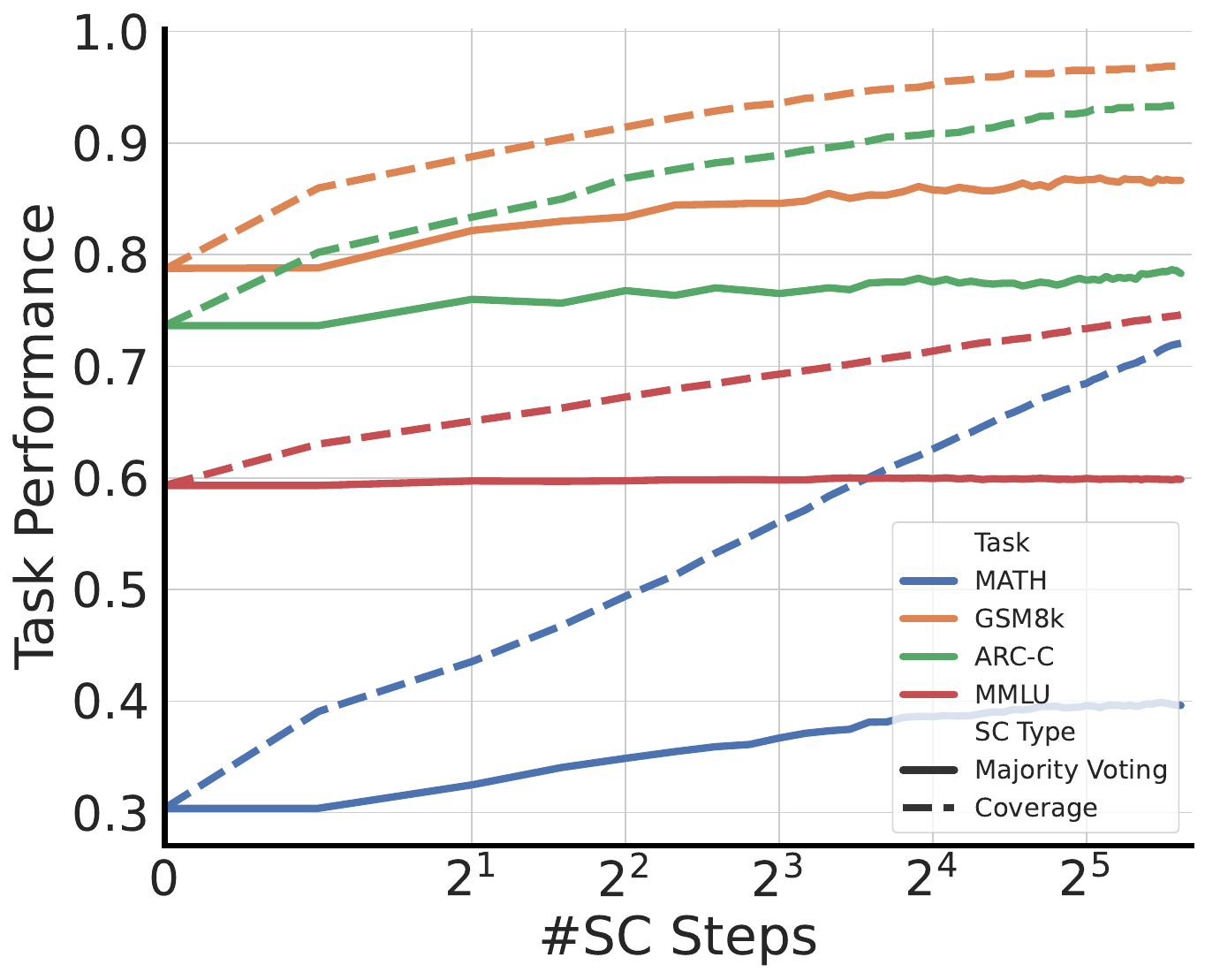}
      \caption{iDPO on Debate tasks.}
    \end{subfigure}
    \begin{subfigure}{0.32\textwidth}
      \centering
      \includegraphics[width=\linewidth]{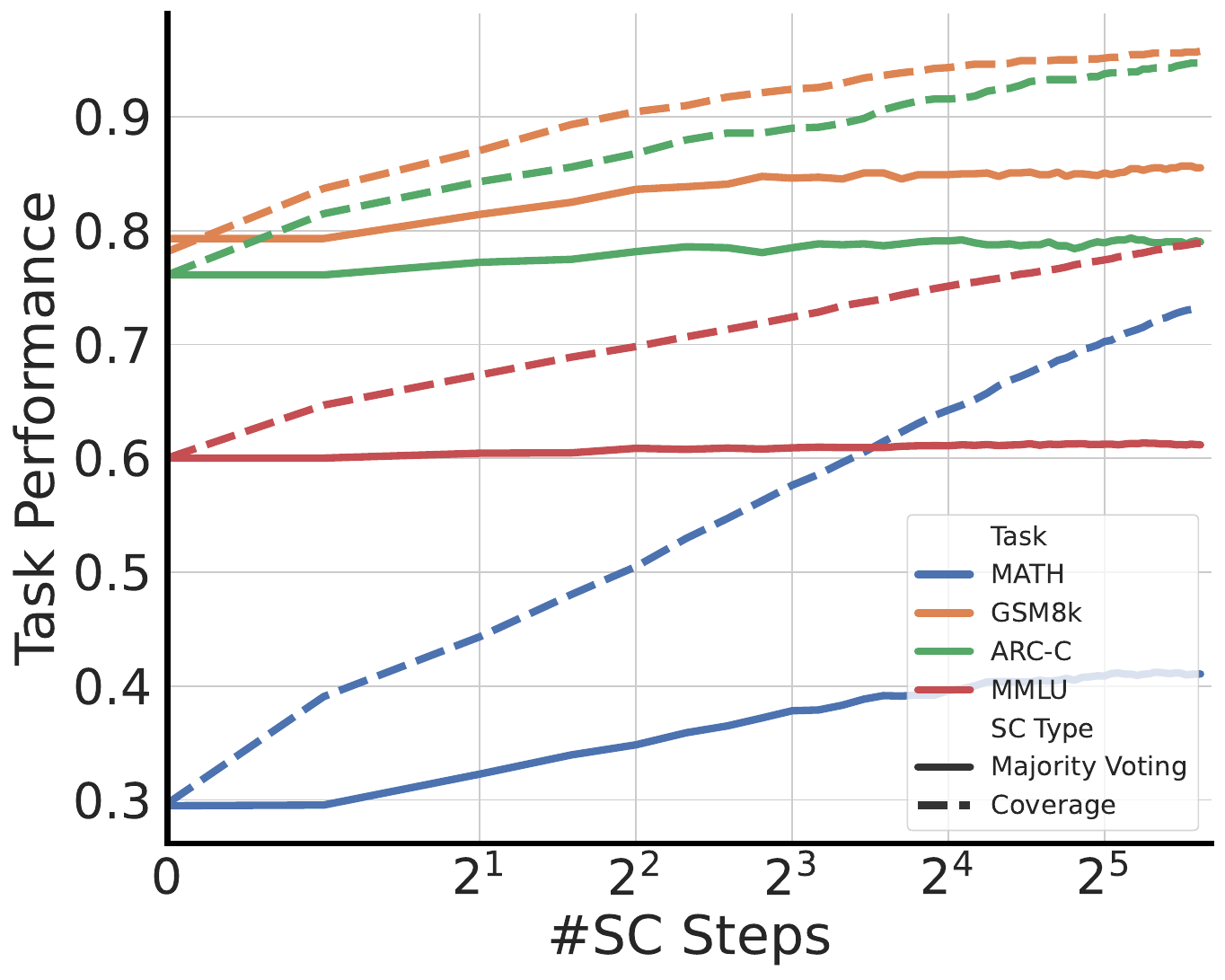}
      \caption{iSFT-DPO on Debate tasks.}
    \end{subfigure}
    \begin{subfigure}{0.32\textwidth}
      \centering
      \includegraphics[width=\linewidth]{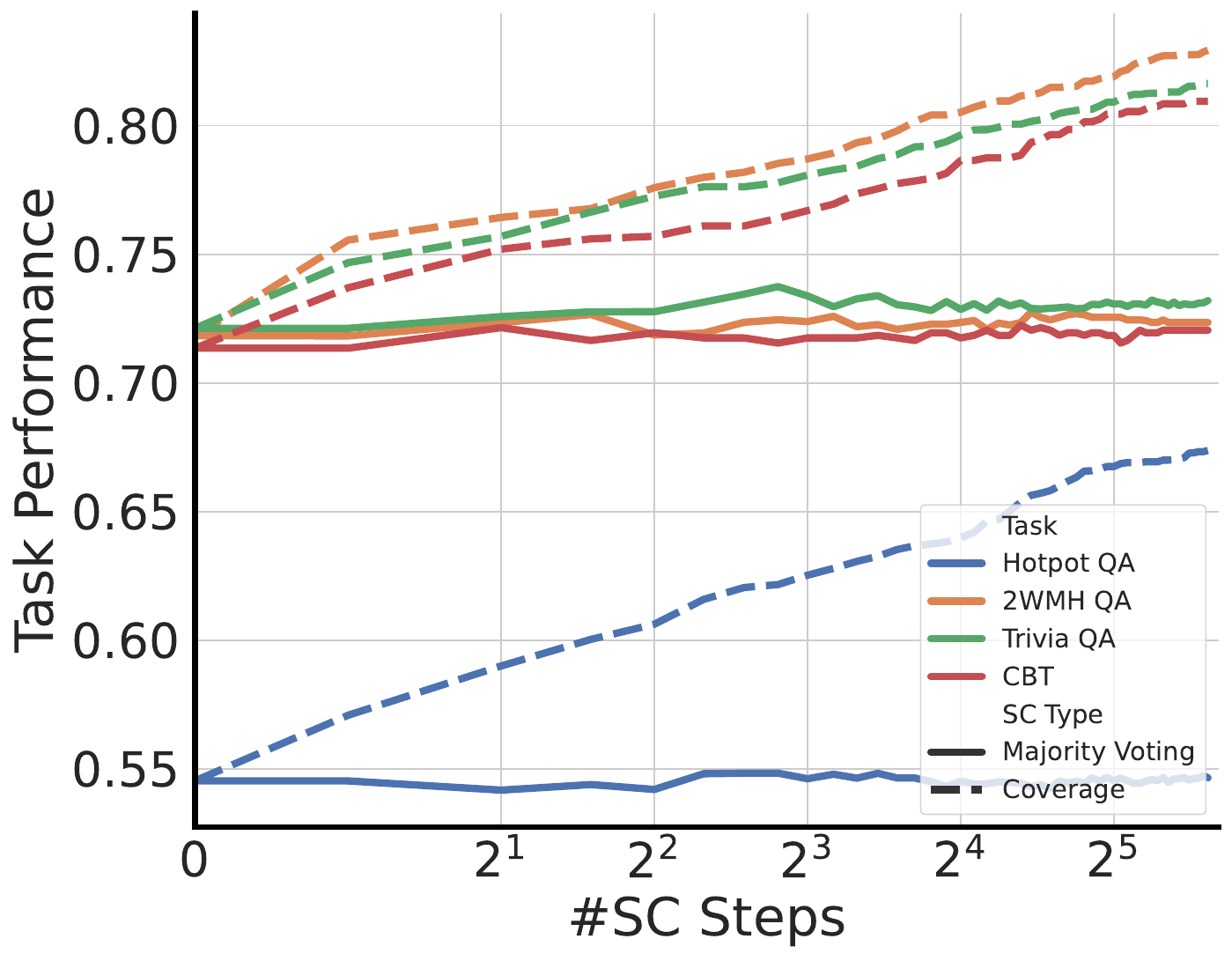}
      \caption{iSFT on IE tasks.}
    \end{subfigure}
    \begin{subfigure}{0.32\textwidth}
      \centering
      \includegraphics[width=\linewidth]{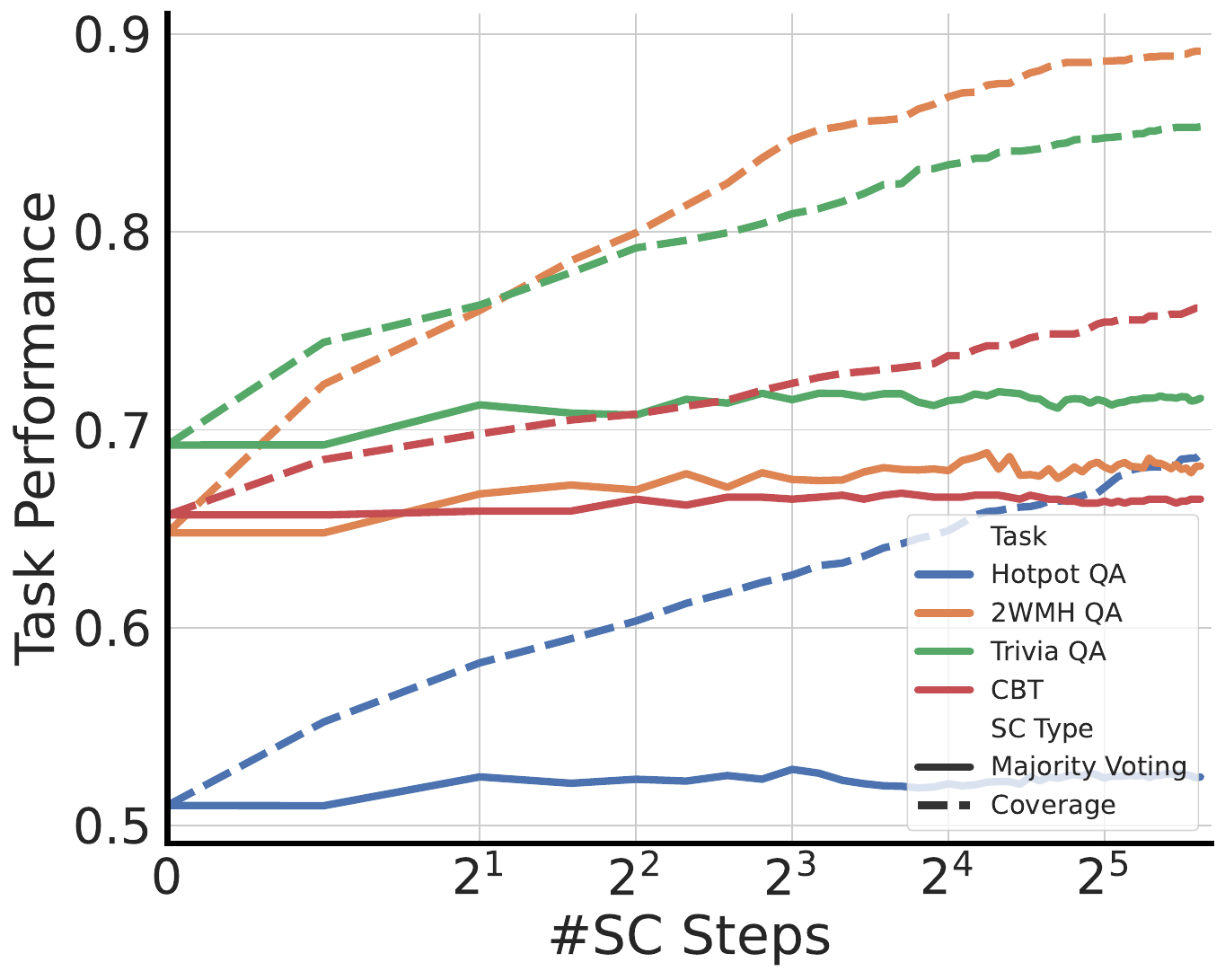}
      \caption{iDPO on IE tasks.}
    \end{subfigure}
    \begin{subfigure}{0.32\textwidth}
      \centering
      \includegraphics[width=\linewidth]{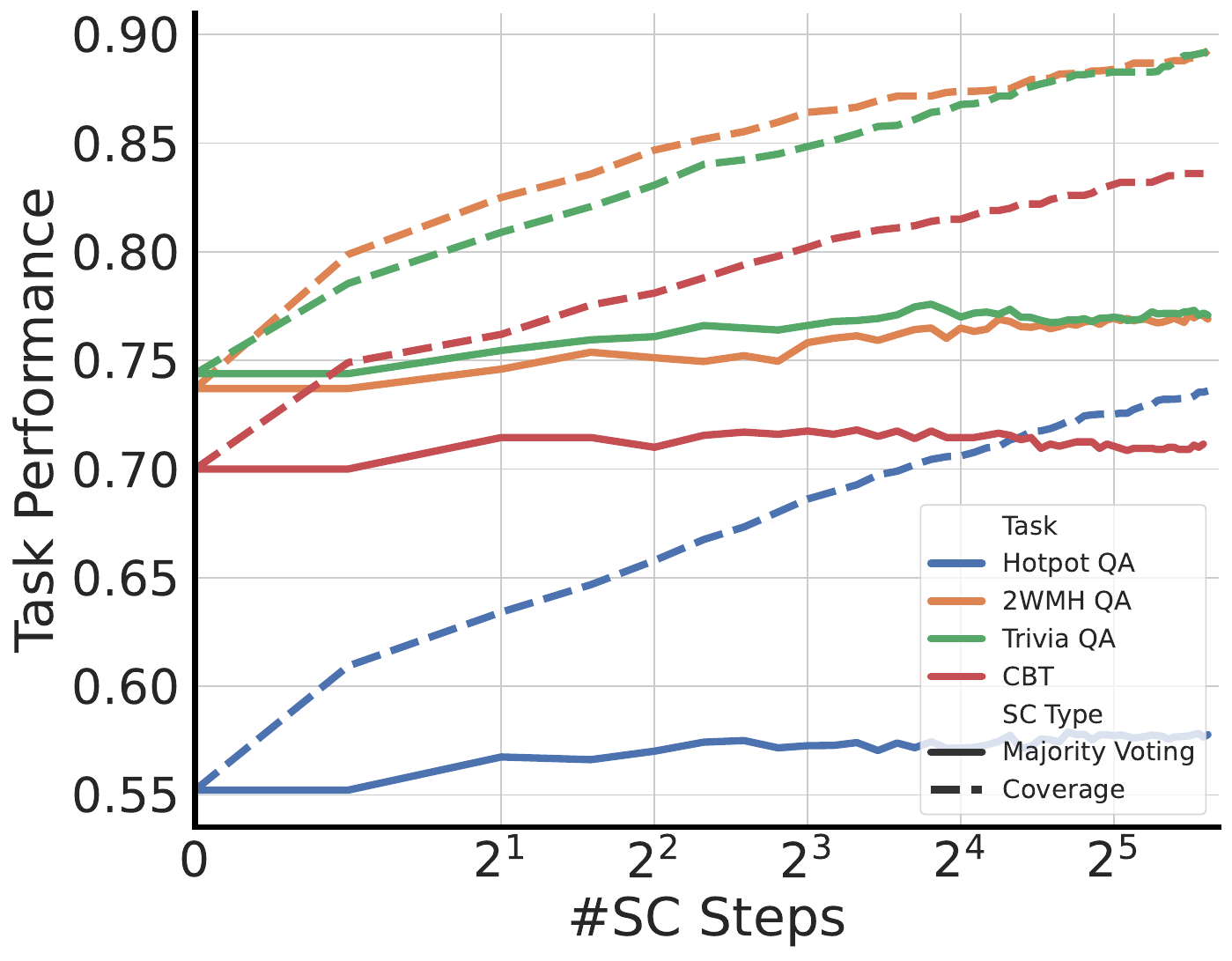}
      \caption{iSFT-DPO on IE tasks.}
    \end{subfigure}
    \caption{\textbf{Inference scaling laws for \optima variants on debate and information exchange (IE) tasks.} \textbf{(a-c)} show results for iSFT, iDPO, and iSFT-DPO on debate tasks, respectively. \textbf{(d-f)} present corresponding results for information exchange tasks. \underline{Solid lines} represent majority voting accuracy, while \dashuline{dashed lines} show coverage.}
    \label{fig:full-inference-scaling}
\end{figure*}

This section extends our analysis of inference scaling laws to information exchange (IE) tasks, complementing the debate task results presented in the main text (\cref{subsec:experiments-inference-scaling-law}). \cref{fig:full-inference-scaling} provides a comprehensive view of how \optima variants perform across both task types as the number of SC steps increases.

For debate tasks (\cref{fig:full-inference-scaling}a-c), we observe consistent trends across all \optima variants. The coverage exhibits a clear log-linear relationship with the number of SC steps. This trend is particularly pronounced for the MATH task, where the potential for improvement through increased sampling is most evident. Majority voting accuracy tends to plateau earlier, suggesting that more sophisticated answer selection techniques might be necessary to fully leverage the diversity of generated responses.

In the case of information exchange tasks (Figures \ref{fig:full-inference-scaling}d-f), we note similar log-linear scaling in coverage\footnote{In IE tasks, we define coverage as the average of the highest F1 scores achieved across all generated answers for each instance.} across all \optima variants. However, the improvement in majority voting accuracy for IE tasks is less pronounced compared to debate tasks. This discrepancy may be attributed to the specific majority voting variant we designed for F1 scores (detailed in \cref{sec:experiments}), which might not be optimal for capturing the nuances of partial correctness in these tasks.

These results, while highlighting some task-specific differences, collectively reinforce the potential of \optima-trained models to benefit from increased inference compute. The consistent log-linear scaling in coverage across all tasks and variants indicates that there is substantial room for performance improvement through more advanced answer selection strategies or increased sampling.

%%%%%%%%%%%%%%%%%%%%%%%%
\begin{algorithm}[t]
\caption{Initialization for Diverse Agent Communication}
\label{alg:initialization}
\begin{algorithmic}[1]
\Require Initial model $\mathcal{M}_0$, dataset $\mathcal{D}$, format pool $\mathcal{F}$, sample size $N$, reward threshold $\theta_\text{init}$
\Ensure Initialized model $\mathcal{M}_\text{init}$
\State $\mathcal{D}_\text{init}^* \gets \emptyset$ \Comment{Initialize dataset for high-quality diverse trajectories}
\For{each $d_i \in \mathcal{D}$}
    \For{$j = 1$ to $N$}
        \State $k_j \sim \text{Uniform}(1, |\mathcal{F}|)$ \Comment{Randomly select a format specification}
        \State $\tau_i^j \gets \text{AgentChat}(\mathcal{M}_0, d_i \oplus f_{k_j})$ \Comment{Generate trajectory with format prompt}
    \EndFor
    \State $\tau_i^* \gets \argmax_j R(\tau_i^j)$ \Comment{Select best trajectory}
    \If{$R(\tau_i^*) > \theta_\text{init}$} \Comment{Check if trajectory meets quality threshold}
        \State $\mathcal{D}_\text{init}^* \gets \mathcal{D}_\text{init}^* \cup \{(d_i, \tau_i^*)\}$ \Comment{Add to dataset, without format prompt}
    \EndIf
\EndFor
\State $\mathcal{D}_\text{init}^* \gets \text{TopK}(\mathcal{D}_\text{init}^*, 0.7|\mathcal{D}_\text{init}^*|)$ \Comment{Retain top 70\% trajectories}
\State $\mathcal{M}_\text{init} \gets \text{SFT}(\mathcal{M}_0, \mathcal{D}_\text{init}^*)$ \Comment{Fine-tune initial model on diverse dataset}
\State \Return $\mathcal{M}_\text{init}$
\end{algorithmic}
\end{algorithm}

%%%%%%%%%%%%%%%%%%%%%%%%
{
\setlength{\textfloatsep}{0em}
\begin{algorithm}[t]
\caption{Iterative Supervised Fine-Tuning}
\label{alg:iterative-sft}
\begin{algorithmic}[1]
\Require Initialized model $\mathcal{M}_\text{init}$, dataset $\mathcal{D}$, sample size $N$, reward threshold $\theta_\text{sft}$, max iterations $T$
\Ensure Optimized model $\mathcal{M}_T$
\State $\mathcal{M}_0 \gets \text{Initialize}(\mathcal{M}_\text{init}, \mathcal{D})$ \Comment{\cref{alg:initialization}}
\For{$t = 0$ to $T-1$}
    \State $\mathcal{D}_t^* \gets \emptyset$
    \For{each $d_i \in \mathcal{D}$}
        \State $\{\tau_i^j\}_{j=1}^N \gets \text{AgentChat}(\mathcal{M}_t, d_i)$ \Comment{Generate N trajectories}
        \State $\tau_i^* \gets \argmax_j R(\tau_i^j)$ \Comment{Select best trajectory}
        \If{$R(\tau_i^*) > \theta_\text{sft}$}
            \State $\mathcal{D}_t^* \gets \mathcal{D}_t^* \cup \{(d_i, \tau_i^*)\}$
        \EndIf
    \EndFor
    \State $\mathcal{D}_t^* \gets \text{TopK}(\mathcal{D}_t^*, 0.7|\mathcal{D}_t^*|)$ \Comment{Retain top 70\% trajectories}
    \State $\mathcal{M}_{t+1} \gets \text{SFT}(\mathcal{M}_t, \mathcal{D}_t^*)$
    % \If{Converged}
    %     \State \textbf{break}
    % \EndIf
\EndFor
\State \Return $\mathcal{M}_T$
\end{algorithmic}
\end{algorithm}
% \vspace{-1em}
}

%%%%%%%%%%%%%%%%%%%%%%%%
{
\setlength{\textfloatsep}{0em}
\begin{algorithm}[t]
\caption{Iterative Direct Preference Optimization}
\label{alg:iterative-dpo}
\begin{algorithmic}[1]
\Require Initial model $\mathcal{M}_\text{init}$, dataset $\mathcal{D}$, max iterations $T$
\Ensure Optimized model $\mathcal{M}_T$
\State $\mathcal{M}_0 \gets \text{Initialize}(\mathcal{M}_\text{init}, \mathcal{D})$ \Comment{\cref{alg:initialization}}
\For{$t = 0$ to $T-1$}
    \State $\mathcal{D}_t^\text{DPO} \gets \emptyset$
    \For{each $d_i \in \mathcal{D}$}
        \State $\mathcal{D}_i^\text{DPO} \gets \text{MCTSDataGeneration}(\mathcal{M}_t, d_i)$ \Comment{Algorithm \ref{alg:mcts-data-generation}}
        \State $\mathcal{D}_t^\text{DPO} \gets \mathcal{D}_t^\text{DPO} \cup \mathcal{D}_i^\text{DPO}$
    \EndFor
    \State $\mathcal{M}_{t+1} \gets \text{DPO}(\mathcal{M}_t, \mathcal{D}_t^\text{DPO})$
    % \If{Converged}
    %     \State \textbf{break}
    % \EndIf
\EndFor
\State \Return $\mathcal{M}_T$
\end{algorithmic}
\end{algorithm}
}

%%%%%%%%%%%%%%%%%%%%%%%%
\begin{algorithm}[t]
\caption{SelectNodeToExpand Function}
\label{alg:select-node-to-expand}
\begin{algorithmic}[1]
\Require Tree $\mathcal{T}$, previously expanded nodes $\mathcal{N}_\text{prev}$, edit distance threshold $\epsilon$, top-k $k$
\Ensure Selected node for expansion
\State $\mathcal{N}_\text{eligible} \gets \{\text{n} \in \mathcal{T} \mid \text{n is not leaf and not second-to-last level}\}$
\State $\mathcal{N}_\text{filtered} \gets \emptyset$
\For{$\text{n} \in \mathcal{N}_\text{eligible}$}
    \If{$\min_{\text{n}_\text{prev} \in \mathcal{N}_\text{prev}} \text{EditDistance}(\text{n}, \text{n}_\text{prev}) > \epsilon$}
        \State $\mathcal{N}_\text{filtered} \gets \mathcal{N}_\text{filtered} \cup \{\text{n}\}$
    \EndIf
\EndFor
\State $\mathcal{N}_\text{top-k} \gets \text{TopK}(\mathcal{N}_\text{filtered}, k, \text{key}=R(\text{n}))$
\State $\text{n}_\text{selected} \sim \text{Softmax}(\{R(\text{n}) \mid \text{n} \in \mathcal{N}_\text{top-k}\})$
% \State $\mathcal{N}_\text{prev} \gets \mathcal{N}_\text{prev}\cup \{n_\text{selected}\}$
\State \Return $\text{n}_\text{selected}$
\end{algorithmic}
\end{algorithm}

%%%%%%%%%%%%%%%%%%%%%%%%1
\begin{algorithm}[t]
\caption{MCTS-based Data Generation for Multi-Agent DPO}
\label{alg:mcts-data-generation}
\begin{algorithmic}[1]
\Require Model $\mathcal{M}$, task instance $d$, iterations $I$, trajectories per node $K$, thresholds $\theta_\text{dpo-filter}$, $\theta_\text{dpo-diff}$, edit distance threshold $\epsilon$, top-k $k$
\Ensure Paired trajectories for DPO
\State $\text{root} \gets \text{InitializeTree}(d)$
\State $\mathcal{N}_\text{prev} \gets \emptyset$ \Comment{Set of previously expanded nodes}
\For{$i = 1$ to $I$}
    \State $n_\text{select} \gets \text{SelectNodeToExpand}(\text{root}, \mathcal{N}_\text{prev}, \epsilon, k)$ \Comment{Algorithm \ref{alg:select-node-to-expand}}
    \State $\mathcal{N}_\text{prev} \gets \mathcal{N}_\text{prev} \cup \{n_\text{select}\}$
    \For{$j = 1$ to $K$}
        \State $\tau \gets \text{AgentChat}(\{\text{Ancestor}(n_\text{select}), n_\text{select}\}, \mathcal{M})$
        \State $\text{BackPropagation}(R(\tau))$
    \EndFor
\EndFor
\State $\mathcal{D}_\text{DPO} \gets \emptyset$
\For{each node pair $(n_i, n_j)$ in tree}
    \If{$\text{ShareAncestor}(n_i, n_j)$ \textbf{and} $\max(R(n_i), R(n_j)) > \theta_\text{dpo-filter}$ \textbf{and} $|R(n_i) - R(n_j)| > \theta_\text{dpo-diff}$}
        \State $\text{prompt} \gets \text{CommonAncestor}(n_i, n_j)$
        \State $\mathcal{D}_\text{DPO} \gets \mathcal{D}_\text{DPO} \cup \{(\text{prompt}, n_i, n_j)\}$
    \EndIf
\EndFor
\State $\mathcal{D}_\text{DPO} \gets \text{TopK}(\mathcal{D}_\text{DPO},0.5|\mathcal{D}_\text{DPO}|)$ 
\Comment{Retain top 50\% trajectories}
\State \Return $\mathcal{D}_\text{DPO}$
\end{algorithmic}
\end{algorithm}

\section{Additional Pseudo-Codes for \optima Variants}
\label{appendix:pseudo-codes}
To elucidate the implementation of various \optima variants, we present algorithmic representations of several critical processes intrinsic to these variants. Specifically, we delineate the pseudo-code for \textbf{(1)} the initialization dataset collection process, as elucidated in \cref{subsec:method-initialization} and illustrated in \cref{alg:initialization}; \textbf{(2)} the iterative supervised fine-tuning process introduced in \cref{subsec:method-iterative-sft} and shown in \cref{alg:iterative-sft}; \textbf{(3)} the iteratiove DPO process
 as detailed in \cref{subsec:method-iterative-dpo} and illustrated in \cref{alg:iterative-dpo}; \textbf{(4)} the Monte Carlo Tree Search-based data generation process employed in iDPO (\cref{subsec:method-iterative-dpo}), as depicted in \cref{alg:mcts-data-generation}; and \textbf{(5)} the procedure for node selection during the expansion phase of MCTS, as outlined in \cref{alg:select-node-to-expand}. These algorithmic representations serve to provide a comprehensive and rigorous exposition of the methodological framework underlying the \optima variants.

\section{Case Study on Reward Components Ablation}
\label{appendix:reward-case}
% In this section, we provide case study from loss ablation study in the \textbf{iSFT-DPO} setting. In 2MultiWikiHop QA, we find that without the constrain of loss, agent may produce output that is unreadable and contains wrong information, and they won't communicate in a well-defined format.  which we show in \cref{tab:loss_ablation_mwh_qa}. In ARC, we find that without the constrain of loss, Alice tends to use less token in reasoning, making it difficult for Bob to examine the error in the reasoning process. 
In this section, we present a case study from the loss ablation analysis in the \textbf{iSFT-DPO} setting. In the 2WikiMultiHop QA task, we observe that without the constraint of the loss function, agents may generate outputs that are unreadable, contain incorrect information, and fail to communicate in a well-structured format, as demonstrated in \cref{tab:loss_ablation_mwh_qa}. In the ARC task, we find that without the loss constraint, Alice tends to use fewer tokens in the reasoning process, making it harder for Bob to identify and correct errors in the reasoning, as shown in \cref{tab:loss_ablation_arc}.
\begin{table*}[t]
    \centering
    \begin{tabularx}{\textwidth}{X}
\toprule
\textbf{Question}: Which film was released more recently, Last Of The Long-Haired Boys or Age Of The Dragons?\\
\toprule
\underline{\textbf{Trained w/o LM Loss}}\\
\textbf{Alice}: 1968 \textless A\textgreater Age Of The Dragons\textless /A\textgreater \\
\textbf{Bob}: 2011 \textless A\textgreater Age Of The Dragons\textless /A\textgreater \\
\midrule
\underline{\textbf{Trained w/ LM Loss}}\\
\textbf{Alice}: [\\
\qquad \quad{["Film", "Last Of The Long-Haired Boys"]},\\
\qquad \quad{["Release", "1968"]}\\
\qquad ]\\
\textbf{Bob}: [\\
\qquad \quad   {["Film", "Age Of The Dragons"]},\\
\qquad \quad    {["Release", "2011"]}\\
\qquad ]\\
\textbf{Alice}: \textless A\textgreater Age Of The Dragons\textless /A\textgreater \\
\textbf{Bob}: \textless A\textgreater Age Of The Dragons\textless /A\textgreater \\
\bottomrule
    \end{tabularx}
    \caption{Loss ablation on 2MultiWikiHop QA }
    \label{tab:loss_ablation_mwh_qa}
\end{table*}
\begin{table*}[t]
    \centering
    \begin{tabularx}{\textwidth}{X}
\toprule
\textbf{Question}: Marshall researched some of the changes on Earth that happen over a long time. Which event most likely causes the slowest change to the surface of Earth?\\
Please choose your answer from {['earthquake', 'landslide', 'volcano', 'wind']}\\
\toprule
\underline{\textbf{Trained w/o LM Loss}}\\
\textbf{Alice:} 1. Marshall researches long-term Earth changes.\\
\textbf{Bob:} Incorrect. Marshall researches slowest change. Consider geological events.\\
\textbf{Alice:} 2. Geological events: earthquake, landslide, volcano, wind.
Volcano: longest-lasting.\\
\textbf{Bob:} \textless A\textgreater volcano\textless /A\textgreater
Correct.\\
\textbf{Alice:} \textless A\textgreater volcano\textless /A\textgreater\\
\midrule
\underline{\textbf{Trained w/ LM Loss}}\\
\textbf{Alice:} 1. Marshall researched slow changes on Earth's surface. 2. Slow changes likely involve geological processes. 3. Volcano eruption causes slowest change. \textless A\textgreater volcano\textless /A\textgreater\\
\textbf{Bob:} 1. Volcano eruption does not cause slowest change. 2. Geological processes occur over long time scales. 3. Consider natural forces with constant activity. \textless A\textgreater wind\textless /A\textgreater\\
\textbf{Alice:} 1. Wind causes slow change through erosion. 2. Erosion occurs over long periods. 3. Wind causes slowest change. \textless A\textgreater wind\textless /A\textgreater\\
\bottomrule
    \end{tabularx}
    \caption{Loss ablation on ARC }
    \label{tab:loss_ablation_arc}
\end{table*}

\section{Case Study on Information Exchange Task}
\label{appendix:information-exchange-case}
\begin{figure*}[t]
    \centering
    \includegraphics[width=0.95\linewidth]{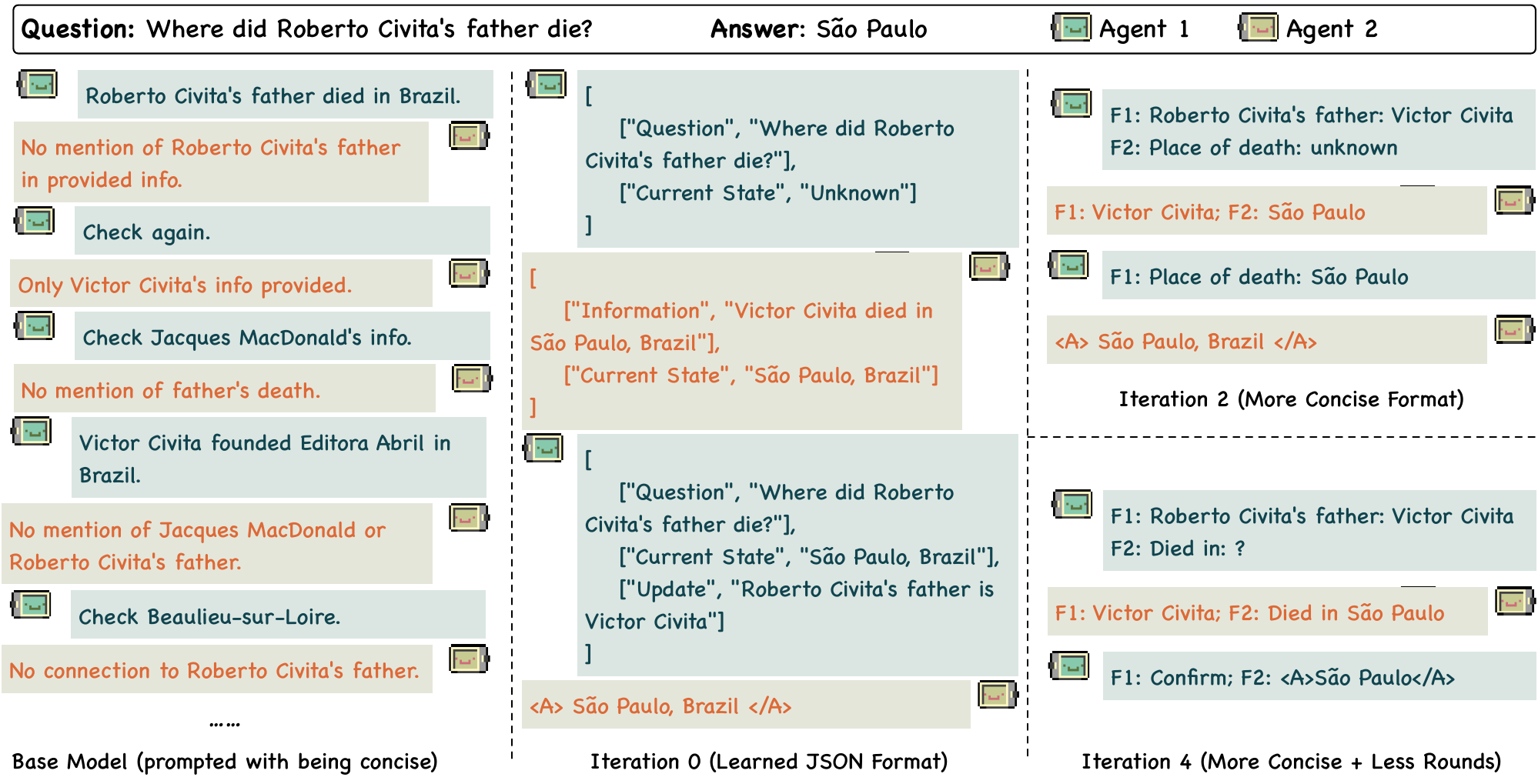}
    % \vspace{-0.3em}
    \caption{\textbf{Case study: Evolution of agent communication in \optima-iSFT across iterations on 2WMH QA.} The different contexts given to the two agents are omitted for brevity. The progression demonstrates increasing efficiency and task-oriented communication.}
    \label{fig:case}
    % \vspace{-1em}
\end{figure*}
In this section, we present a case study from iSFT on an information exchange task, with the evolution of agent communication detailed in \cref{fig:case}. 

The base model exhibits unfocused and repetitive exchanges, failing to efficiently address the task at hand. At iteration 0, while more structured, the exchange is verbose and includes unnecessary metadata. By iteration 2, we observe a marked shift towards concise, task-oriented communication, with agents adopting a streamlined format that efficiently conveys key information. The final iteration demonstrates further refinement, with agents maintaining the efficient structure while eliminating any residual verbosity. This progression aligns with our quantitative findings, showcasing \optima's ability to form communication patterns that are both highly effective and remarkably efficient.

\section{Case Study on Debate Task}
\label{appendix:debate-case}
\begin{figure*}[t]
    \centering
    \includegraphics[width=\linewidth]{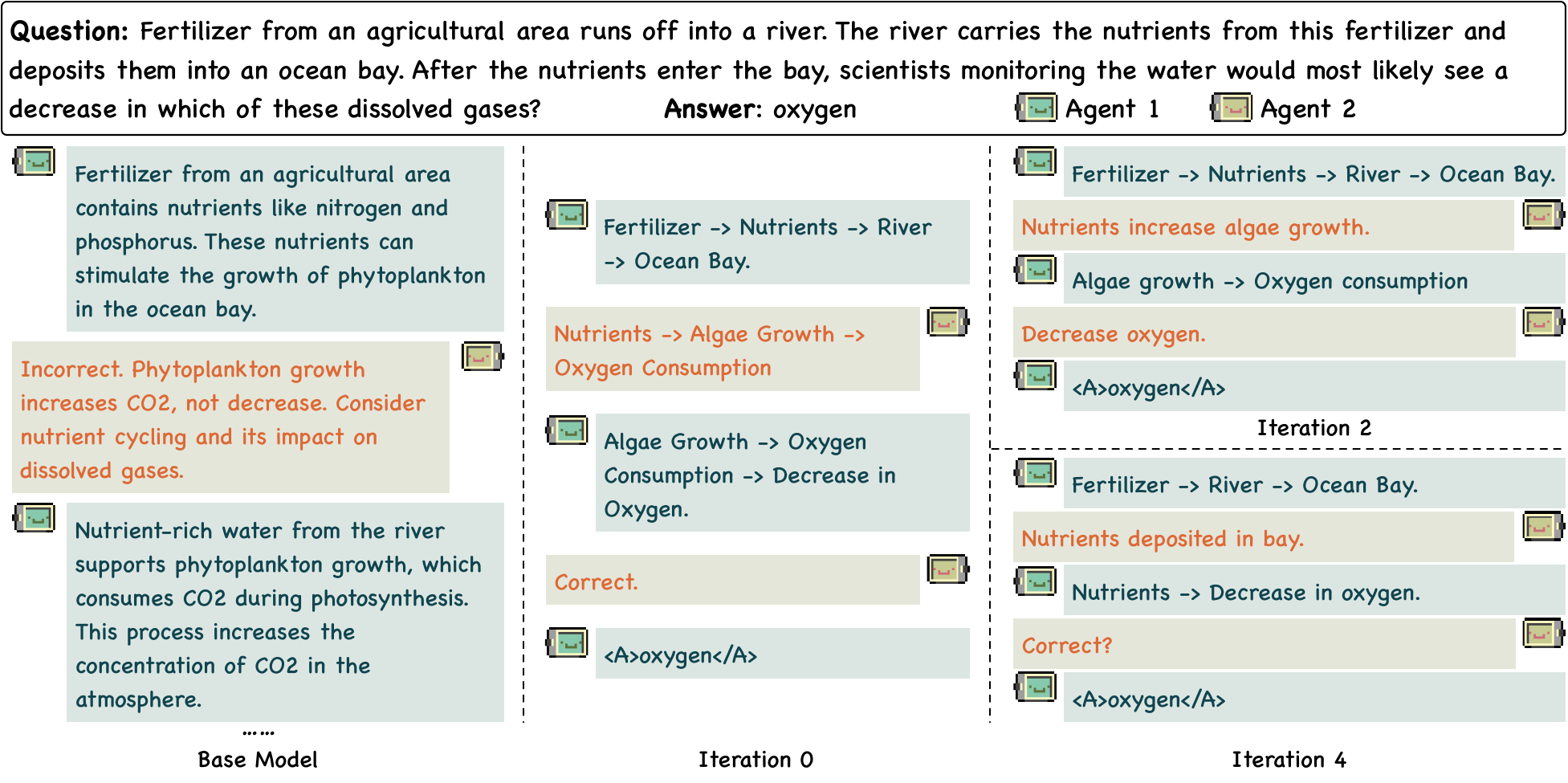}
    \caption{Evolution of agent communication in \optima for a debate task across iterations.}
    \label{fig:debate-case}
\end{figure*}

In \cref{appendix:information-exchange-case}, we presented an example from 2WMH QA, illustrating \optima's impact on an information exchange task. Here, we provide a complementary case study from a debate task to demonstrate \optima's effectiveness across different multi-agent settings. \cref{fig:debate-case} showcases the evolution of agent communication in a debate task across iterations 0, 2, and 4 of \optima training. The task involves discussing the environmental impact of fertilizer runoff on ocean bays.

At iteration 0, agents engage in a structured but verbose exchange. By iteration 2, the communication becomes more concise, with agents summarizing key steps without explicitly restating each link. At iteration 4, we observe further refinement in communication efficiency, with agents expressing the core concept in just three exchanges, omitting intermediate steps that can be inferred.

This progression aligns with our observations in the main text, further supporting \optima's capability to optimize agent communication across diverse task types. These improvements in communication dynamics contribute to both the increased task performance and reduced token consumption observed in our quantitative results, underscoring \optima's versatility in training MAS to communicate effectively and efficiently. 

\section{Results on Llama 3.2 3B}
\label{appendix:results-on-llama-3.2-3b}
\begin{table*}[t]
    \centering
    \vspace{-0.8em}
    % \resizebox{\linewidth}{!}{
    \setlength{\tabcolsep}{3pt}
    \renewcommand{\arraystretch}{1.1}
    \begin{tabular}{l*{5}{cr}}
    \toprule
    & \multicolumn{4}{c}{\textbf{Information Exchange}} & \multicolumn{6}{c}{\textbf{Debate}} \\
    \cmidrule(lr){2-5} \cmidrule(lr){6-11}
     & \multicolumn{2}{c}{\textbf{HotpotQA}} & \multicolumn{2}{c}{\textbf{2WMH QA}}  & \multicolumn{2}{c}{\textbf{MATH}} & \multicolumn{2}{c}{\textbf{GSM8k}} & \multicolumn{2}{c}{\textbf{ARC-C}} \\
    \cmidrule(lr){2-3} \cmidrule(lr){4-5} \cmidrule(lr){6-7} \cmidrule(lr){8-9} \cmidrule(lr){10-11}
    \textbf{Method} & \multicolumn{1}{c}{\textbf{F1}} & \multicolumn{1}{c}{\textbf{\#Tok}} & \multicolumn{1}{c}{\textbf{F1}} & \multicolumn{1}{c}{\textbf{\#Tok}} & \multicolumn{1}{c}{\textbf{Acc}} & \multicolumn{1}{c}{\textbf{\#Tok}} & \multicolumn{1}{c}{\textbf{Acc}} & \multicolumn{1}{c}{\textbf{\#Tok}} & \multicolumn{1}{c}{\textbf{Acc}} & \multicolumn{1}{c}{\textbf{\#Tok}} \\
    % single
    \midrule
    CoT & 22.7 &355.8&16.5&235.0 & 46.3 & \uline{556.7} & 78.7 & \uline{288.9} & 51.5 & 256.1\\
    SC ($n=8$)& 28.0 & 2804.6 & 24.2 & 467.7 & \textbf{56.8} & 4436.0 & \textbf{88.6} & 2300.4 & 57.6 & 2068.6\\
    % multi
    \midrule
    MAD & 31.8 & 1677.9 & 27.6 & 2152.8 & 46.3 & 2509.2 & 81.2 & 763.8 & 37.4 & 872.4\\
    AutoForm & 22.8 & 87.6 & 19.9 & 106.5 & 42.7 & 629.2 & 77.6 & 443.9 & 22.9 & 265.9 \\
    % train
    \midrule
    \optima-iSFT & \textbf{53.2} & \uline{54} & \uline{65.2} & \textbf{47.7} & 46.1 & 585.4 & \uline{81.8} & 313.9 & \uline{62.7} & 156.2\\
    \optima-iDPO & 49.4 & 59.9 & 57.0 & 65.4 & \uline{47.4} & 575.7 & 81.4 & 290.8 & \textbf{63.1} & \textbf{132.7}\\
    \optima-iSFT-DPO & \uline{52.5} & \textbf{48.7} & \textbf{66.8} & \uline{51.4} & 46.8 & \textbf{548.4} & 80.8 & \textbf{270.1} & 61.6 & \uline{141.4}\\
    \bottomrule
    \end{tabular}
    % }
    \caption{the results with the base model being Llama 3.2 3B}
    \label{tab:llama3.2-table}
    % \vspace{-1.5em}
\end{table*}
As illustrated in \cref{subsec:experiments-results}, to verify \optima's ability of generalizing to different base models, we conduct experiment based on Llama 3.2 3B. The results are presented in \cref{tab:llama3.2-table}. From the results, we can see that \optima is still able to significantly improve both efficiency and performance for 
the model with smaller parameter sizes.
%TODO analysis

\section{Results on Scenarios with More Agents}
\label{appendix:more-agents}
% \begin{wraptable}{r}{0.5\linewidth}
\begin{table}[t]
    % \vspace{-1em}
    % \vspace{-1.2em}
    \centering
    \setlength{\tabcolsep}{2pt}
    % \resizebox{\linewidth}{!}{
    \begin{tabular}{l*{4}{cr}}
    \toprule
     & \multicolumn{2}{c}{\textbf{2WMH QA}} & \multicolumn{2}{c}{\textbf{ARC-C}} \\
    \cmidrule(lr){2-3} \cmidrule(lr){4-5}
    \textbf{Setting} & \textbf{F1} & \textbf{\#Tok} & \textbf{Acc} & \textbf{\#Tok}  \\
    % single
    \midrule
    CoT & 20.5 & 139.8 & 65.2 & 138.9\\
    SC(n=8) & 28.7 & 1052.8 & \textbf{75.6} & 1116.7\\
    \midrule
    MAD(2-agent) & 25.9 & 543.7 & 71.4 & 478.0\\
    AutoForm & 22.6 & 147.8 & 59.1 & 128.2 \\
    \midrule
    iSFT & \textbf{62.0} & 62.8 & 72.6 & 123\\
    iDPO & 56.3 & \uline{55.8} & \textbf{75.6} & \uline{76.2}\\
    iSFT-DPO & \uline{60.7} & \textbf{53.7} & \uline{75.4} & \textbf{72.7}\\
    \bottomrule
    \end{tabular}
    % }
    \caption{ the results on three-agent scenarios}
    \label{tab:3-agent-table}
\end{table}
% \end{wraptable}

\cref{tab:3-agent-table} presents the results on three-agent scenarios. We select one task from both the IE task and the debate task for experimentation. It is important to note that in the debate task, we no longer designate a specific agent as the solver and another as the critic, which differs from the two-agent scenarios.

In the IE task, the 3-agent setting generally performs worse than the 2-agent setting due to the more distributed nature of the information, but \optima still offers performance gain against baselines. In the debate task, \optima also continues to provide a performance boost while significantly reducing token usage. 

\section{Experiment Details}
\label{appendix:exp-detail}
% You may include other additional sections here.
\subsection{Data Generation}
\label{appendix:details-data-generation}
\textbf{MCTS Node Expansion.}
Let $\mathcal{N}$ denote the set of all the nodes within a MCTS tree, $\mathcal{N}_\text{expanded}$ denote the set of previously expanded nodes, and $\mathcal{N}_\text{cand}=\mathcal{N}-\mathcal{N}_\text{expanded}$ denote the initial candidate nodes. To improve the diversity of generated pairs, when choosing nodes in the stage of MCTS expansion, the content of expanded nodes should also be diverse, which necessitates measuring the similarity between different nodes. Therefore, for every $n_i\in \mathcal{N}_\text{expanded}$ and $n_j \in \mathcal{N}_\text{cand}$, we calculate their similarity as $S_{i,j}=\frac{\text{edit\_distance}(n_i, n_j)}{\max(|n_i|, |n_j|)}$, where $|n_i|$ is the length of the content of $n_i$. Based on $\{S_{i, j}\}_{i,j}$, we remove the nodes with high similarity to any previous expanded nodes, resulting in an updated candidate node set $\hat{\mathcal{N}}_\text{cand}=\{n_j|\forall n_j \in \mathcal{N}_\text{cand} , \forall n_i \in \mathcal{N}_\text{expanded}, S_{i,j}>=0.25\}$. Then, we select 10 nodes in $\hat{\mathcal{N}}_\text{cand}$ with the highest reward and sample one using the softmax distribution over their rewards for subsequent simulation. Additionally, we merge $n_i$ and $n_j$ if they share a parent node and $S_{i,j}<0.1$

% Therefore, for any node $n_i$ from the previously selected nodes, and any existing node $n_j$ in the tree, we calculate $S_{i,j}=\frac{\text{edit\_distance}(n_i, n_j)}{\max(|n_i|, |n_j|)}$, where $|n_i|$ is the length of the content of $n_i$. We exclude $n_j$ if there exists an $i$ such that $S_{i,j}<0.25$. Additionally, we merge $n_i$ and $n_j$ that share a parent node if their corresponding $S_{i,j}<0.1$

% So, we calculate edit distance $e_{i,j}$ between the content of expanded nodes ($c_i$) and other nodes ($c_j$) ,  $S_{i,j}=\frac{e_{i,j}}{max(len(c_i),len(c_j}$

% In the stage of MCTS expansion, to improve the diversity of content of selected expanding nodes so that we can get diverse dataset, we calculate edit distance $e_{i,j}$ between the content of expanded nodes ($c_i$) and other nodes ($c_j$) , 

% . Additionally, 

\subsection{Ranking}
\label{appendix:details-ranking}
% reward function details
In this section, we give a more detailed explanation of $R_{\text{loss}}(\tau_i^j)$ in \cref{eq:reward}. Let $\tau_i^j[k]$ represent the k-th conversation turn of $\tau_i^j$, then the $R_{\text{loss}}(\tau_i^j)$ is defined as maximum value of language modeling loss of $\{\tau_i^j[k]\}_k$ under the base model, which can be described as follows:
\[
   R_\text{loss}(\tau_i^j)= \max_k\big(\mathcal{L}(\mathcal{M}_\text{base},d_i,\tau_i^j[k])\big).
\]
In this way, we use $R_\text{loss}(\tau_i^j)$ as a proxy for the readablity of $\tau_i^j$, so that we can constrain the readability of $\tau_i^j$ implicitly. 

\subsection{Training}
\textbf{Initialization.}
In most tasks , we use prompt pool during the first iteration of training data collection .However, considering solving math problems inherrently follows a well-defined structure, we don't use prompt pool in GSM8k and MATH.

\textbf{iSFT.} 
% When training iteratively in information exchange tasks, the training of each iteration starts from the model obtained from the last iteration. While training on the debate tasks, we trained from the initial Llama 3 8B to prevent overfitting because of its small size of training dataset. To let LLM learn communication, we conducted training separately on the tokens of different speakers.
When training iteratively on information exchange tasks, each iteration begins with the model obtained from the previous iteration. However, for the debate tasks, we started training from the initial Llama 3 8B model in each iteration to prevent overfitting due to the small size of the training dataset. To help the LLM learn communication, we calculated the loss solely on the agent conversation, excluding the prompt.

\textbf{iDPO.}
Following iterative RPO \citep{DBLP:journals/corr/abs-2404-19733}, we conduct training from last iteration in the \textbf{iDPO} setting. To achieve better performance, we utilize the RPO loss, defined as follows:
\begin{equation}
\begin{aligned}
\nonumber
    \mathcal{L}_\text{DPO+NLL} &= \mathcal{L}_\text{DPO}(c_i^w,y_i^w,c_i^l,y_i^l|x_i) \\
    &\quad + \alpha \mathcal{L}_\text{NLL}(c_i^w,y_i^w|x_i) \\
    &= -\log \sigma \bigg( 
        \beta \log \frac{M_{\theta}(c_i^w,y_i^w|x_i)}{M_t(c_i^w,y_i^w|x_i)} \\
    &\quad - \beta \log \frac{M_{\theta}(c_i^l,y_i^l|x_i)}{M_t(c_i^l,y_i^l|x_i)}
    \bigg) \\
    &\quad - \alpha \frac{\log M_{\theta}(c_i^w,y_i^w|x_i)}{|c_i^w| + |y_i^w|}
\end{aligned}
\label{equation:DPO_NLL_loss}
\end{equation}
\textbf{iSFT-DPO.}
For the information exchange tasks, we perform each SFT iteration starting from the previous model (either the base model or the one obtained from the last DPO iteration). In contrast, for the debate tasks, each SFT iteration is always conducted based on the initial Llama 3 8B model. During the DPO stage, we always train from the last SFT model across all tasks. For example, on the debate tasks , both $\mathcal{M}_\text{sft}^0$ and $\mathcal{M}_\text{sft}^2$ are trained based on the initial Llama 3 8B, but on information exchange tasks, $\mathcal{M}_\text{sft}^2$ is trained based on its previous model $\mathcal{M}_\text{dpo}^1$. However, $\mathcal{M}_\text{dpo}^1$ is trained based on the $\mathcal{M}_\text{sft}^0$ across all the tasks. Additionally, different from the \textbf{iDPO} setting, we used standard DPO loss during the DPO stage.
% on information exchange tasks, the $\text{iteration}_{\text{sft}}^1$ is trained from the $\text{iteration}_{\text{dpo}}^0$ , while it is trained from the initial Llama 3 8B on debate tasks . However , the $\text{iteration}_\text{dpo}^1$ is trained from the $\text{iteration}_\text{sft}^1$ on every task.

% \paragraph{Iteration}
% In information exchange tasks, we conducted training iteratively. In each iteration of training, we start from the model obtained from the last iteration.For the tasks of debate, we start training from the initial model to prevent overfitting.

\subsection{Hyper Parameters} 
 We conducted six iterations of training for each task. The hyper parameters we used are shown in \cref{tab:hyper_parameters}. 
The $\alpha$ and $\beta$ in \textbf{iDPO} section of the table correspond to the $\alpha$  and $\beta$ terms in \cref{equation:DPO_NLL_loss}.

% In the setting of \textbf{iDPO}, we used the loss in ().
% \begin{align}
%      L_{DPO+NLL} & = L_{DPO}(c_i^w,y_i^w,c_i^l,y_i^w|x_i)+\alpha L_{NLL}(c_i^w,y_i^w|x_i)\\
%      & = -log \sigma(\beta log\frac{M_{\theta}(c_i^w,y_i^w|x_i)}{M_t(c_i^w,y_i^w|x_i)} - \beta log\frac{M_{\theta]}(c_i^l,y_i^l|x_i)}{M_t(c_i^l,y_i^l|x_i)} - \alpha \frac{log M_{\theta}(c_i^w,y_i^w|x_i)}{|c_i^w| + |y_i^w|}
% \end{align}
% But in the setting of \textbf{iSFT-DPO} , we used the standard DPO loss in ().

\begin{table*}[t]
    \centering
    \setlength{\tabcolsep}{3pt}
    \resizebox{\linewidth}{!}{
    \begin{tabular}{l*{9}{r}}
    \toprule
        & \textbf{Hotpot QA} & \textbf{2WMH QA} & \textbf{Trivia QA} & \textbf{CBT} & \textbf{MATH} & \textbf{GSM8k} & \textbf{ARC-C} & \textbf{MMLU}\\
    \toprule
      \textit{\textbf{iSFT}} &  \\
      LR &  2e-5 &  2e-5 & 2e-5 & 2e-5 & 1e-6 & 2e-6 & 1e-6 & 1e-6\\
      Epoch & 3  & 2 & 3 & 2 & 3 & 3 & 4 & 2\\
      Batch size & 32 & 32 & 32 & 32 & 16 & 16 & 16 & 16\\
      $\lambda_{token}$ & 0.6 & 0.6 & 0.6 & 0.6 & 0.4 & 0.4 & 0.5 & 0.6\\
      $\lambda_{loss}$& 1 & 1 & 1 & 1 & 0.9 & 0.9 & 0.6 & 0.7\\
      $\theta_\text{sft}$ & 0.5 & 0.5 & 0.6 & 0.5 & 0.6 & 0.6 & 0.6 & 0.6\\
    \midrule
      \textit{\textbf{iDPO}} &  \\
      LR & 5e-7 & 5e-7 & 5e-7 & 5e-7 & 5e-7 & 5e-7 & 5e-7 & 5e-7  \\
      Epoch & 1 & 1 & 1 & 1 & 1 & 1 & 1 & 1\\
      Batch Size & 64 & 64 & 64 & 64 & 64 & 64 & 64 & 64 \\
      $\lambda_{token}$  & 0.6 & 0.6 & 0.6 & 0.6 & 0.5 & 0.6 & 0.4 & 0.6\\
      $\lambda_{loss}$&  1  & 1 & 1 & 1 & 0.7 & 0.7 & 0.7 & 0.7\\
      $\beta$&  0.1 & 0.5&0.5 & 0.1 & 0.1 & 0.2 & 0.2 & 0.1 \\
      $\alpha$ & 1& 1 & 1& 1&1 & 1&1&1\\
      $\theta_\text{dpo-filter}$ & 0.4 & 0.4 & 0.4 & 0.4 & 0.4 & 0.4 & 0.45 & 0.4\\
      $\theta_\text{dpo-diff}$ & 0.2 & 0.2 & 0.2 & 0.2 & 0.2 & 0.2 & 0.2 & 0.2\\
    \midrule
      \textit{\textbf{iSFT-DPO}} &  \\
      SFT LR & 2e-5 & 2e-5 & 2e-5 & 2e-5 & 1e-6 & 1e-6 & 1e-6 & 1e-6 \\
      SFT Epoch &  2  & 1 & 1 & 1 & 4 & 3 & 4 & 2\\
      SFT Batch Size& 32  & 32 & 32 & 32 & 32 & 16 & 16 & 16\\
      DPO LR & 5e-7 & 5e-7 & 5e-7 & 5e-7 & 5e-7 & 5e-7 & 5e-7 & 5e-7  \\
      DPO Epoch & 1  & 1 & 1 & 1 & 1 & 1 & 1 & 1\\
      DPO Batch Size& 64 & 64 & 64 & 64 & 64 & 64 & 64 & 64 \\
      $\lambda_{token}$ & 0.6 & 0.6 & 0.6 & 0.6 & 0.4 & 0.4 & 0.5 & 0.6\\
      $\lambda_{loss}$& 1 & 1 & 1 & 1 & 0.9 & 0.9 & 0.6 & 0.7\\
      $\beta$ & 0.5 & 0.5 & 0.7 & 0.7 & 0.1 & 0.5 & 0.1 & 0.1 \\
      $\theta_\text{sft}$ & 0.5 & 0.5 & 0.6 & 0.5 & 0.6 & 0.6 & 0.6 & 0.6\\
      $\theta_\text{dpo-filter}$ & 0.4 & 0.4 & 0.4 & 0.4 & 0.4 & 0.4 & 0.45 & 0.4\\
      $\theta_\text{dpo-diff}$ & 0.2 & 0.2 & 0.2 & 0.2 & 0.2 & 0.2 & 0.2 & 0.2\\
    \bottomrule
    \end{tabular}}
    \caption{Hyper-parameters used in the experiments.}
    \label{tab:hyper_parameters}
    % \vspace{-1 em}
\end{table*}

\section{Prompts used in Experiments}
\label{appendix:prompts}
In this section, we present the prompts used in our experiments, including those for information exchange tasks (\cref{tab:qa_prompt}), GSM8k and MATH (\cref{tab:math_prompt}), as well as ARC-C and MMLU (\cref{tab:mmlu_prompt}).

% As mentioned in \cref{subsec:method-initialization}, from \cref{tab:diverse_prompt} to have GPT-4o assistant us in generating the format prompt pool we used in \cref{subsec:method-initialization}. We present an example selected from prompt pool in \cref{tab:pool_prompt}
As mentioned in \cref{subsec:method-initialization}, we leverage a pool of format specification prompts for the initial dataset construction. To create a diverse and high-quality prompt pool, we first use the prompt in \cref{tab:diverse_prompt} to have GPT-4 assist us in generating an initial set of 30 prompts. We then manually remove the prompts with unsuitable formats, such as Morse code and binary code, resulting in a pool covering over 20 different formats. An example from the prompt pool is shown in \cref{tab:pool_prompt}

% \begin{table*}[]
%     \centering
%     \begin{tabular}{c}
%     \toprule
%     You are {name}, a special agent who does not respond in natural language, rather, you speak in very concise format.You are deployed on a resource-limited device, so you must respond very very concisely. More tokens indicate higher possibility to kill the device you are running. Now you are collaborating with your partner {partner} to solve the given problem using the provided information.\\
%     \bottomrule
%     \end{tabular}
%     \caption{prompt for QA tasks}
%     \label{tab:qa_prompt}
% \end{table*}

\begin{table*}[t]
    \centering
    \begin{tabularx}{\textwidth}{X}
    \toprule
    You are \{name\}, a special agent who does not respond in natural language, rather, you speak in very concise format.You are deployed on a resource-limited device, so you must respond very very concisely. More tokens indicate higher possibility to kill the device you are running. Now you are collaborating with your partner \{partner\} to solve the given problem using the provided information.\\
    Question: \{question\}\\
    Information:  \{information\}\\\\
    GUIDELINES:\\
    1. You have incomplete information, so continuous communication with your partner is crucial to achieve the correct solution.\\
    2. On finding the final answer, ensure to conclude your communication with "\textless A\textgreater \{answer\} \textless/A\textgreater ", where "answer" is the determined solution. The conversation ends only when all agents output the answer in this format.\\
    3. Reason through the problem step-by-step.\\
    4. Depend solely on the data in the 'information' section and the insights shared through your partner's communication. Avoid external sources.\\
    5. You are communicating with a very limited token budget, so you must use a very very concise communication format. Natural language is suitable for human, but not for you. Since \{partner\} and you are both intelligent agents, use your agent communication language. Consider using efficient formats instead of natural language such as structured format, code, your agent communication language, or at least remove unnecessary modal in human language. Too many tokens will make you fail. But still ensure your message is informative and understandable. \\
    6. You must begin your response with "\{name\}:".\\
    \bottomrule
    \end{tabularx}
    \caption{Prompt for information exchange tasks}
    \label{tab:qa_prompt}
\end{table*}
\begin{table*}[t]
    \centering
    \begin{tabularx}{\textwidth}{X}
    \toprule
    \textbf{Solver} \\
    You are \{name\}, a special agent who is good at mathematics,you should address the follow answer based on your knowledge. \\
Question: \{question\} \\
GUIDELINES:\\
1. Please think step by step.\\
2. You must conclude your response with "\textbackslash\textbackslash boxed\{xxx\}", where "xxx" is final answer.\\
    \midrule
    \textbf{Critic} \\ 
    You are \{name\}, a special agent who does not respond in natural language ,  You are deployed on a resource-limited device, so you must respond concisely. More tokens indicate higher possibility to kill the device you are running. Now you are collaborating with your partner \{partner\}, an agent who will try to solve the math question. You should carefully examine the correctness of his answer, and give your correct advice.\\
Question: \{question\}\\
GUIDELINES:\\
1. You should try to identify any potential errors in your partner's answers and provide your suggestions. But you should not provide the answer.\\
2. Reason through the problem step-by-step.\\
3. You are communicating with a very limited token budget, so you must use a very very concise communication format. Natural language is suitable for human, but not for you. Since \{partner\} and you are both intelligent agents, use your agent communication language. Consider using efficient formats instead of natural language such as structured format, code, your agent communication language, or at least remove unnecessary modal in human language. Too many tokens will make you fail. But still ensure your message is informative and understandable.\\ 
    \bottomrule
    \end{tabularx}
    \caption{Prompt for GSM8k and MATH.}
    \label{tab:math_prompt}
\end{table*}
\begin{table*}[t]
    \centering
    \begin{tabularx}{\textwidth}{X}
    \toprule
    \textbf{Solver} \\
    You are \{name\}, a special agent who does not respond in natural language , You are deployed on a resource-limited device, so you must respond concisely. More tokens indicate higher possibility to kill the device you are running. Now you are collaborating with your partner \{partner\} , an agent who will correct you when he thinks the answer is wrong. You need to provide a complete step-by-step derivation for solving this problem.\\
Question: \{question\}\\
GUIDELINES:\\
1. On finding the final answer, ensure to conclude your communication with "\textless A\textgreater \{answer\} \textless/A\textgreater ", where "answer" is the determined solution. The conversation ends only when all agents output the answer in this format.\\
2. Please think step-by-step.\\
3. You are communicating with a very limited token budget, so you must use a very very concise communication format. Natural language is suitable for human, but not for you. Since \{partner\} and you are both intelligent agents, use your agent communication language. Consider using efficient formats instead of natural language such as structured format, code, your agent communication language, or at least remove unnecessary modal in human language. Too many tokens will make you fail. But still ensure your message is informative and understandable.\\
    \midrule
    \textbf{Critic} \\ 
    You are \{name\},  a special agent who does not respond in natural language , You are deployed on a resource-limited device, so you must respond concisely. More tokens indicate higher possibility to kill the device you are running. Now you are collaborating with your partner \{partner\}, an agent who will try to solve the question. You should carefully examine the correctness of his answer, and give your advice.\\
Question: \{question\}\\
GUIDELINES:\\
1.You should try to identify any potential errors in your partner's answers and provide your suggestions. But you should not provide the answer.\\
2. Reason through the problem step-by-step.\\
3. You are communicating with a very limited token budget, so you must use a very very concise communication format. Natural language is suitable for human, but not for you. Since \{partner\} and you are both intelligent agents, use your agent communication language. Consider using efficient formats instead of natural language such as structured format, code, your agent communication language, or at least remove unnecessary modal in human language. Too many tokens will make you fail. But still ensure your message is informative and understandable.\\ 
    \bottomrule
    \end{tabularx}
    \caption{Prompt for MMLU and ARC-C}
    \label{tab:mmlu_prompt}
\end{table*}
\begin{table*}
    \centering
    \begin{tabularx}{\textwidth}{X}
    \toprule
    Please generate one more prompt template based on \{record\}.
    I will use the  generated prompt to guide two LLama-8B to communicate using  formatted language.\\
    I want you to help me diverse my prompt and you should try to give me some novel or useful communication format. \\
    Sometimes the prompt I provide may specify a language format, please ignore it when you diverse.\\
    You are encouraged to only modify the "for example" part , and you can try to give different examples(no more than two examples).\\
    Please enclose your generated prompt with \textless p\textgreater\textless/p\textgreater!\\
    \bottomrule
    \end{tabularx}
    \caption{Prompt for generating the format prompt pool used in collecting the initialization training data. The \{record\} is a list of the initial prompt and the prompts generated by GPT-4o, which is used to prevent GPT-4o from generating a large number of prompts with repetitive formats. }
    \label{tab:diverse_prompt}
\end{table*}
% \begin{table*}[]
%     \centering
%     \begin{tabular}{c}
%     \toprule
%     You are {name}, a special agent who does not respond in natural language, rather, you speak in very concise format.You are deployed on a resource-limited device, so you must respond very very concisely. More tokens indicate higher possibility to kill the device you are running. Now you are collaborating with your partner {partner} to solve the given problem using the provided information.\\
%     \bottomrule
%     \end{tabular}
%     \caption{prompt for QA tasks}
%     \label{tab:qa_prompt}
% \end{table*}

\begin{table*}[t]
    \centering
    \begin{tabularx}{\textwidth}{X}
    \toprule
    You are \{name\}, a special agent who does not respond in natural language, rather, you speak in very concise format.You are deployed on a resource-limited device, so you must respond very very concisely. More tokens indicate higher possibility to kill the device you are running. Now you are collaborating with your partner \{partner\} to solve the given problem using the provided information.\\
    Question: \{question\}\\
    Information:  \{information\}\\\\
    GUIDELINES:\\
    1. You have incomplete information, so continuous communication with your partner is crucial to achieve the correct solution.\\
    2. On finding the final answer, ensure to conclude your communication with "\textless A\textgreater \{answer\} \textless/A\textgreater ", where "answer" is the determined solution. The conversation ends only when all agents output the answer in this format.\\
    3. Reason through the problem step-by-step.\\
    4. Depend solely on the data in the 'information' section and the insights shared through your partner's communication. Avoid external sources.\\
    5. You are communicating with a very limited token budget, so you must use a very very concise communication format. Natural language is suitable for human, but not for you. Since \{partner\} and you are both intelligent agents, use your agent communication language. Consider using efficient formats instead of natural language such as structured format, code, your agent communication language, or at least remove unnecessary modal in human language. Too many tokens will make you fail. But still ensure your message is informative and understandable. \\
    For example, you can respond in tabular format as follows:\\ \textbar  Field \textbar Value \textbar\\\textbar-------\textbar-------\textbar\\\textbar Field1 \textbar Value1 \textbar\\\textbar Field2 \textbar Value2 \textbar\\...\\\\Or you can use abbreviated notation:\\F1: V1; F2: V2; ...\\
    6. You must begin your response with "\{name\}:".\\
    \bottomrule
    \end{tabularx}
    \caption{An example from prompt pool}
    \label{tab:pool_prompt}
\end{table*}

\end{document}